\documentclass[10pt,twocolumn]{article} 
\usepackage{simpleConference}
\usepackage{times}
\usepackage{graphicx}
\usepackage{amssymb}
\usepackage{url,hyperref}
\usepackage{multirow}
\usepackage{subcaption}
\usepackage{url}
\usepackage{natbib}

\begin{document}

\title{Where to look at the movies : Analyzing visual attention to understand movie editing}

\author{Alexandre Bruckert, Marc Christie, Olivier Le~Meur \\
\\
Univ. Rennes, CNRS, IRISA, France \\
\\
alexandre.bruckert@irisa.fr  \\
}

\maketitle
\thispagestyle{empty}

\begin{abstract}
In the process of making a movie, directors constantly care about where the spectator will look on the screen. Shot composition, framing, camera movements or editing are tools commonly used to direct attention. In order to provide a quantitative analysis of the relationship between those tools and gaze patterns, we propose a new eye-tracking database, containing gaze pattern information on movie sequences, as well as editing annotations, and we show how state-of-the-art computational saliency techniques behave on this dataset. In this work, we expose strong links between movie editing and spectators scanpaths, and open several leads on how the knowledge of editing information could improve human visual attention modeling for cinematic content. The dataset generated and analysed during the current study is available at \url{https://github.com/abruckert/eye_tracking_filmmaking} 

\end{abstract}

\section{Introduction}
\label{sec:intro}
In order to deal with the incredibly large amount of data coming from our visual environment, human beings have developed a biological mechanism called overt visual attention. While watching a scene, the eye makes sudden noncontinuous movements called saccades, only stopping during events called fixations. Eye fixations occur so that regions of visual interest are centered on the densest zone in photoreceptors of the retina, called the fovea. This area is heavily packed with cone cells, which allows maximum visual acuity, even if it only represents approximately one degree of the visual field. Several studies have shown that eye fixations and visual attention are closely associated~\citep{Findlay1997}. Therefore, studying gaze patterns is of great interest in a wide range of fields~\citep{Duchowski2002, Zhang2020}. For instance, \cite{Karessli2017} showed that human gaze is class discriminative, and thus can help improve classification models. In image and video processing, visual attention and saliency have been widely used in compression algorithms~\citep{Yu2009,Zund2013,Hadizadeh2014}. In the medical field, eye-tracking devices are used to help the communication in cases of locked-in syndrom~\citep{Majaranta2002} or for diagnosis purposes~\citep{Harezlak2016}; for more applications in medicine, see for instance~\citet{Harezlak2018}. 

The factors explaining where people look in a video are usually divided into two categories: \textit{bottom-up} and \textit{top-down} factors. Top-down characteristics refer to observer dependant properties, such as the age of the observers, their cultural background, or the task at hand. These factors have been shown to be the cause of sometimes extreme discrepancies in gaze patterns; see for instance~\citet{LeMeur2017} for an exploration of the age factor, or~\citet{Chua2005} and~\citet{Rayner2009} for the cultural parameter. Bottom-up factors refer to stimuli characteristics, such as the spatial properties of the visual scene, or the temporal characteristics of a video. It also includes the implicit properties of the stimuli, such as the presence of faces~\citep{Cerf2008} or text in the scene. Most of the visual attention models are purely bottom-up models, meaning that they only extract information from the stimulus. Indeed, bottom-up visual saliency has proven to be a reliable predictor of fixations location in images~\citep{Borji2013}.

Over the last century, filmmakers have developed an instinctive knowledge of how to guide the gaze of the audience, manipulating bottom-up characteristics, such as visual cuts, camera movements, shot composition and sizing, and so on. This empirical knowledge contributes to building a set of cinematographic rules and conventions, designed to accommodate the artistic intention of the director with the perception of the audience. However, formalizing these visual tools is not an easy task, and several frameworks and languages have been proposed~\citep{Ronfard2013,Wu2017,Wu2018}. Such languages help quantifying common cinematographic rules, and allow automated models to be used in the movie production process.

As a consequence, studying the quantitative perceptual effects of the visual tools available to filmmakers is of great interest, both for understanding the way humans perceive movies, but also for filmmakers themselves, who could get quantitative feedback on the effects of their work and techniques. Understanding the mechanisms underlying the visual attention on movies can also be of help for computational models related to movie production, such as automated camera placement, automated editing or 3D animated scenes design.

In this paper, we extend the work of \citet{Breeden2017} by proposing a new eye-tracking data\-base on 20 different movie clips, of duration 2 to 7 minutes each. For each clip, we provide cinematographic features annotations drawn from~\citet{Wu2017}, such as the camera movement and angle, the framing size, and the temporal location of cuts and edits. Alongside with a comprehensive analysis of the collected data, we expose several strong correlations between high-level cinematographic features and gaze patterns, that can be easily used to improve visual attention modeling. We also perform a benchmark of visual attention models on this database, and we show that state-of-the-art models often struggle to grasp and use these high-level cinematographic characteristics. Finally, we discuss several leads on how that information could be included in human visual attention models, in order to improve their performances on cinematic content. 

\section{Related work}
\label{sec:related_work}

In this section, we provide a quick overview of the recent works in visual attention modeling, and especially bottom-up approaches, such as visual saliency modeling. We then give a very brief review of the field of visual attention in the context of cinematography, and of the databases available to conduct such studies.

\subsection{Modeling visual attention}
As mentioned earlier, eye movements rely on two kinds of attention mechanisms: top down (or endogenous) influences, which are shaped by high-level cognitive processes, such as the task at hand, the cultural background of the observer, or its medical condition, and bottom-up (or exogenous) movements, which are driven by the features of the stimulus itself. The most common way of representing attention, whether it is endogenous or exogenous, is through a representation called \textit{saliency map}, which is a distribution predicting the likelihood of an eye fixation to occur at a given location. In this work, we will mostly focus on this representation, even if it is not the only one, nor does it captures the full range of human visual attention mechanisms~\citep{Foulsham2008, Koehler2014}. 

There has been only a few studies dedicated to mo\-del top-down visual attention in scenes. For instance, \citet{Kanan2009} proposed a top-down saliency detector, based on object appearance in a Bayesian framework. Other attempts of such models, by~\citet{Jodogne2007TopDown} or~\citet{Borji2011TopDown} for example, yield decent predictive results, considering that the internal cognitive state of an observer is extremely hard to predict, and can lead to less coordination and congruency among gaze patterns of observers~\citep{Mital2011, Bruckert2019}.

On the other hand, many attention models are dealing with bottom-up features (see for instance~\citet{Borji2013Review, Borji2019Review, Wang2018Review} for extensive reviews). Early models focused on static images, using linear filtering to extract meaningful feature vectors, which are then used to predict a saliency map~\citep{Itti98, Bruce2005, LeMeur2006, Harel2006GBVS, Gao2009}. Those meaningful visual features include contrast, orientation, edges, or colors, for instance. In the case of dynamic scene viewing, the early investigations underlined the importance of temporal features, such as optical flow or flicker~\citep{Guo2010, Mahadevan2010, Mital2011, Rudoy2013}. Most of the early dynamic saliency models are however extensions of existing static models, and are limited by the representation power of the chosen hand-crafted features, therefore not grasping the full amount of information delivered by ground-truth saliency.

Recently, deep learning approaches managed to significantly improve performances of attention models. The first attempt of using automatically extracted features was conducted by~\citet{Vig2014}, and managed to outperform most models of the state of the art at the time. Later on, several deep learning models using transfer-learning were proposed, where features learned on large-scale classification datasets were used, like DeepFix~\citep{Kruthiventi2015DeepFix}, SALICON~\citep{Huang2015Salicon}, DeepNet~\citep{Pan2016DeepNet} or Deep~Gaze~II~\citep{Kummerer2017DeepGazeII}. More recently, the emergence of large-scale fixation datasets allowed for end-to-end approaches, in order to learn features more specific to visual saliency. These new models, like SalGan~\citep{Pan2017SalGAN}, SAM-VGG and SAM-Resnet \citep{Cornia2018SAM}, or MSI-Net~\citep{Kroner2020MSINet}, exhibit great predictive behaviors, and constitute a very strong baseline for modeling human visual attention. Dynamic models followed the same path towards deep learning, with models such as DeepVS~\citep{Jiang2018DeepVS}, ACLNet~\citep{Wang2018Review}, \cite{Bak2018DynamicSaliency} or~\cite{Gorji2018}. Similarly to the static case, they exhibit significantly better predictive performan\-ces than earlier approaches (see for instance~\citet{Wang2019review} for a more detailed review).

\subsection{Visual attention and movies}
Studying film perception and comprehension is still an emerging field, relying on broader studies on scene perception~\citep{Smith2012Review, Smith2013Review}. While the effects of low-level features have been studied in great detail, in part thanks to the progress of saliency models, the effects of higher-level film characteristics are far less well understood. \citet{Loschky2014Jaws} showed that the context of a sequence is particularly relevant to understand the way humans are viewing a particular shot, thus underlying the need for a better comprehension of the high-level features. \citet{Valuch2015} studied the influence of colors during editorial cuts, showing that continuity editing techniques result in faster re-orientations of gaze after a cut, and that color contributes to directing attention during edits. Other studies showed strong relationships between eye movement patterns and the number and the size of faces in a scene~\citep{Rahman2014, Cutting2016}.

A few studies focused on gaze congruency, or attentional synchrony. \citet{Goldstein2007} showed that observers tend to exhibit very similar gaze patterns while watching films, and that the inter-observer agreement would be sufficient for effective attention based applications, like magnification around the most important points of the scene. \citet{Mital2011, Smith2013Synchrony} later showed that attentional synchrony was positively correlated with low-level features, like contrast, motion and flicker. Breathnach~\cite{Breathnach2016AttentionalSA} also studied the effect of repetitive viewing on gaze agreement, showing a diminution of the inter-observer congruency when movie clips were watched several times.

More generally, it appears that understanding human visual attention while watching movies ultimately requires a framework combining both low- and high-level features. From a cognitive point of view, \citet{Loschky2020} recently proposed a perception and comprehension theory, distinguishing between the front-end processes, occurring during a single fixation, and back-end processes, occurring across multiple fixations and allowing a global understanding of the scene. From a computational and modeling point of view, no model combining low- and high-level film characteristics has yet been proposed. Alongside with \citet{Breeden2017}, this paper aims to facilitate the development of such a model.

\subsection{Movies eye-tracking datasets}
\label{sec:datasets}
In the field of visual attention modeling for videos, a majority of the large-scale databases used to train various models contain mostly non-cinematographic stimuli. As we show in Section~\ref{sec:visual_attention_modeling}, this leads to consistent errors when saliency models are used on film sequences. Moreover, most studies involving visual attention and movies use their own collected eye-tracking data, as the experimental setups are often very specific to the characteristics studied. However, there exists a few available eye-tracking databases on movie scenes, that can be general enough for modeling purposes.

\textbf{Hollywood-2}~\citep{Mathe2015Hollywood2} includes 1707 movie clips, from 69 Hollywood movies, as well as fixation data on those clips from 19 observers. Observers were split into three groups, each with a different task (3 observers free-viewing, 12 observers with an action recognition task, and 4 observers with a context recognition task). Each group being relatively small, the common way to use this data for visual attention modeling is by merging those groups, thus introducing potential biases. The large scale of this dataset (around 20 hours of video) is well fit for training deep saliency models, however few conclusions regarding ga\-ze patterns on movies can be drawn from the data itself, since it mainly focuses on task-driven viewing mode, and that each clip is only around 15 seconds long.

\textbf{SAVAM}~\citep{Gitman2014Savam} includes 41 high-definition videos, 28 of which are movie sequences (or use movie-like realisation, like commercials for instan\-ce). Eye fixations are recorded from 50 observers, in a free viewing situation. As for Hollywood-2, the each clip is quite short, only 20 seconds on average.

\textbf{Breeden and Hanrahan} \citeyearpar{Breeden2017} proposed eye-tra\-cking data from 21 observers, on 15 clips from 13 films, for a total of 38 minutes of content. Each clip is between 1 and 4 minutes. Alongside this data, they also provide high-level feature annotations, such as the camera movements in shots, the temporal location and types of edits, the presence or absence of faces on screen, and  whether or not the characters are speaking. However, the main limitations of this dataset are the relatively low precision of the eye-tracking device used, and the duration of the total content of the base itself.

It follows that the saliency modeling community, as well as cinematographic studies, would greatly benefit from an extension of Breeden and Hanrahan's work, \emph{i.e.} a relatively large-scale eye-tracking database on movies sequences, including a large diversity of editing styles, genres and epochs, alongside with high-level features annotations, related to different film-making parameters. In this work, we propose such a database, and the conclusions that we can draw from it.

\section{Dataset overview}
\subsection{Films and clips selection}
In \citet{Wu2017}, the authors proposed a language called \textit{Film Editing Patterns} (FEP) to annotate the production and edition style of a film sequence. Alongside this formalization of cinematographic rules, they present an open database of annotations on several film sequences, for pattern analysis purposes. In order to simplify the annotation process of our dataset, we decided to use the same clips. 

We selected 20 clips, extracted from 17 different movies. The movies span different times (from 1966 to 2012) and genres, and are from different directors and editors, in order to eliminate bias coming from individual style. Table~\ref{tab:MoviesSelection} gives an overview of the selected clips. The sequences were selected as they were the most memorable or famous sequences from each movie, based on scenes that users uploaded to YouTube, indicating popularity and interest to the general public.

\begin{table*}[h]
    \caption{Overview of the selected clips}
    \centering
    \resizebox{\linewidth}{!}{\begin{tabular}{|c|c|c|c|c|c|}
    \hline
        \textbf{Title} & \textbf{Director} & \textbf{Genre (IMDb)} & \textbf{Nb. Frames} & \textbf{Aspect ratio} & \textbf{Year}  \\ \hline
        American History X & Tony Kaye & Drama & 5702 & 1.85 & 1998\\ \hline
        Armageddon & Michael Bay & Action, Adventure, Sci-Fi & 4598 & 2.39 & 1998\\ \hline
        The Curious Case of Benjamin Button & David Fincher & Drama, Fantasy, Romance & 4666 & 2.40 & 2008\\ \hline
        Big Fish & Tim Burton & Adventure, Drama, Fantasy & 3166 & 1.37 & 2003\\ \hline
        The Constant Gardener & Fernando Meirelles & Drama, Mystery, Romance & 5417 & 1.85 & 2005\\ \hline
        Departures & Yôjirô Takita & Drama, Music & 10117 & 1.85  & 2008\\ \hline
        Forrest Gump & Robert Zemekis & Drama, Romance & 2689 & 2.39 & 1994\\ \hline
        Gattaca (1) & Andrew Niccol & Drama, Sci-Fi, Thriller & 3086 & 2.39 & 1997\\ \hline
        Gattaca (2) & Andrew Niccol & Drama, Sci-Fi, Thriller &  3068 & 2.39 & 1997\\ \hline
        The Godfather & Francis Ford Coppola & Crime, Drama & 1918 & 1.37 & 1972\\ \hline
        The Good, The Bad \& The Ugly & Sergio Leone & Western & 9101 & 2.35 & 1966\\ \hline
        The Hunger Games & Gary Ross & Action, Adventure, Sci-Fi & 5771 & 2.35 & 2012\\ \hline
        Invictus & Clint Eastwood & Biography, Drama, History & 2203 & 2.39 & 2009\\ \hline
        LOTR : The Fellowship of the Ring & Peter Jackson & Action, Adventure, Drama & 5109 & 2.40 & 2001\\ \hline
        Pulp Fiction & Quentin Tarantino & Crime, Drama & 3211 & 2.39 & 1994\\ \hline
        The Shawshank Redemption (1) & Frank Darabont & Drama & 5374 & 1.85 & 1994\\ \hline
        The Shawshank Redemption (2) & Frank Darabont & Drama & 4821 & 1.85 & 1994\\ \hline
        The Shining & Stanley Kubrick & Drama, Horror & 4781 & 1.33 & 1980\\ \hline
        The Help (1) & Tate Taylor & Drama & 4151 & 1.85 & 2011\\ \hline
        The Help (2)& Tate Taylor & Drama & 5244 & 1.85 & 2011\\ \hline
    \end{tabular}}
    \label{tab:MoviesSelection}
\end{table*}

Here we give a small description of each scene, and its most remarkable characteristics:
\begin{itemize}
    \item \textbf{American History X}: Flashback scene, dialogue between characters seated at a table. Mostly static shots on the faces of the characters. This scene is in black and white.
    \item \textbf{Armageddon}: Action scene, high frequency of edits. The shot size varies a lot, from extreme closeups to large establishing shots. A lot of camera movements.
    \item \textbf{Benjamin Button}: Flashback scene. A lot of camera movements tracking the characters. A narrator comments the whole sequence. Some of the shots are replicated, with variations, in order to indicate alternative possibilities in the unfolding of the narrated story.
    \item \textbf{Big Fish}: Crowd scene, with two main characters walking through the crowd. A few shots take place in a whole different location, with only the two characters conversing.
    \item \textbf{The Constant Gardener}: Dramatic scene, the camera is handheld, and follows a single character throughout the sequence.
    \item \textbf{Departures} : Closing scene, alternation of static camera shots. Three characters are present, but no dialogue.
    \item \textbf{Forrest Gump}: Flashback scene, narrated by a character. Camera movements are used to reveal actors in the scene.
    \item \textbf{Gattaca (1)}: Dialogue scene between two characters. A lot of play on camera angles, since one of the characters is in a wheelchair, and the other one is standing.
    \item \textbf{Gattaca (2)}: Dialogue scene between three characters.
    \item \textbf{The Godfather} : Dramatic sequence, where the edits alternate back and forth from one central quiet scene to several simultaneous dramatic situations. 
    \item \textbf{The Good, The Bad and The Ugly}: Mexican standoff scene, with three characters, where the frequency of the edits accelerate and the shot sizes go from larger to closer as the tension builds up.
    \item \textbf{The Hunger Games}: Dramatic scene, alternating a lot of different camera movements, angles and shot sizes. A crowd is present, but several tricks (colored clothing, focus) are used to distinguish the main characters.
    \item \textbf{Invictus}: Contemplative scene, starting in a cell and ending in outdoors. Characters appear and disappear as ghosts. A narrator reads a poem.
    \item \textbf{Lord of The Rings}: Dialogue scene between two characters, alternating with flashbacks, mostly of action scenes. Different camera movements, angles and shot sizes.
    \item \textbf{Pulp Fiction}: Dialogue scene between two characters seated face to face. The exact same camera angle is used throughout the scene. 
    \item \textbf{Shawshank Redemption (1)}: Dialogue between several characters, various camera movements, angles and shot sizes.
    \item \textbf{Shawshank Redemption (2)}: Flashback scene, following a single character, explaining a prison escape. A narrator comments a part of the sequence. Various camera movements, angles and shot sizes.
    \item \textbf{The Shining}: Dialogue scene between two characters. Very low frequency of edits, and abundant presence of the color red in the scene.
    \item \textbf{The Help (1)}: Flashback scene, dialogue between two characters.
    \item \textbf{The Help (2)}: Flashback scene, in between a dialogue scene between two characters. A lot of faces and colored clothing.
\end{itemize}

The length of the clips varies from 1 minute 30 to 7minutes. This length is voluntarily higher than in the other datasets presented in Section 2.3, in order to allow the observer to feel immersed in the sequence, and thus exhibiting more natural gaze patterns. In total, the dataset contains roughly one hour of content. Table~\ref{tab:ShotsCharacteristics} show the lengths of the average shots for each sequence. The high diversity in terms of shot lengths underlines the diversity in terms of editing styles.

\begin{table*}[h]
    \caption{Lengths of the sequences, and of the longest, shortest and average shots of each sequence.}
    \resizebox{\linewidth}{!}{\begin{tabular}{|c|c|c|c|c|}
    \hline
        \textbf{Sequence} & \textbf{Sequence Length (s)} & \textbf{Longest shot (s)} & \textbf{Shortest shot (s)} & \textbf{Average shot (s)}  \\ \hline
        Armageddon & 191.8 & 12.1 & 0.0 & 1.6\\ \hline
        The Hunger Games & 240.8 & 16.7 & 0.6 & 2.4\\ \hline
        The Curious Case of Benjamin Button & 194.7 & 11.8 & 0.3 & 2.5\\ \hline
        The Godfather & 80.0 & 6.8 & 0.5 & 2.7\\ \hline
        Big Fish & 132.1 & 7.6 & 0.7 & 2.8\\ \hline
        The Constant Gardener & 226.0 & 13.8 & 0.4 & 3.5\\ \hline
        LOTR : The Fellowship of the Ring & 213.1 & 8.4 & 0.5 & 3.6 \\ \hline
        The Good, The Bad \& The Ugly & 379,7 & 36.5 & 0.2 & 3.8\\ \hline
        The Help (2) & 218.8 & 14.0 & 1.0 & 4.0\\ \hline
        Invictus & 91.9 & 8.6 & 1.8 & 4.2\\ \hline
        American History X & 237.9 & 14.7 & 1.0 & 4.2\\ \hline
        Pulp Fiction & 134.0 & 12.2 & 1.4 & 4.6 \\ \hline
        The Shawshank Redemption (1) & 224.2 & 19.2 & 0.8 & 4.7\\ \hline
        The Help (1) & 173.2 & 17.7 & 1.8 & 6.0\\ \hline
        Gattaca (1) & 128.7 & 23.7 & 0.2 & 6.1\\ \hline
        Departures & 422.0 & 21.8 & 1.8 & 6.6\\ \hline
        Forrest Gump & 112.2 & 16.6 & 1.8 & 6.7\\ \hline
        Gattaca (2) & 128.0 & 17.1 & 1.8 & 6.7\\ \hline
        The Shawshank Redemption (2) & 201.1 & 18.0 & 1.8 & 7.7\\ \hline
        The Shining & 199.5 & 107.1 & 8.8 & 39.9\\ \hline
    \end{tabular}}
    \label{tab:ShotsCharacteristics}
\end{table*}

\subsection{High-level features annotations}
\label{annotations}
Films typically contain many high-level features aiming to attract or to divert the observers' visual attention~\citep{Smith2012Review}. These features can be of different sorts~: the presence of faces or text, the framing properties, the scene composition, or the camera motion and angle, for instance. The timing of the shots, the selection of the shots from rushes by the editor and the narrative it creates are also high-level features specific to films. Audio cues, like the presence of music or dialogue can also be considered as a form of high-level movie features, and have been increasingly studied as a way to improve visual attention models~\citep{Tavakoli2019DAVE}. However, all of those features can prove very challenging to extract automatically, which can explain why saliency models seem to only learn non-temporal image characteristics, at the scale of the frame, like contrast- or texture-like information. We then used the database of Film Editing Patterns described in \cite{Wu2018} to select a hand-crafted set of high-level annotations that can help in the study of visual attention and gaze patterns on films. More particularly, such annotations enable us to conduct quantitative analysis on the influence of these cinematographic features over visual attention.

\subsubsection{Camera motion}

Camera motion is an efficient tool used on set by the filmmaker to direct attention. For each shot of the data\-base, we differentiate several possible camera motions:
\begin{itemize}
    \item \textit{Static}: The camera is mounted on a stand and does not move.
    \item \textit{Track}: The camera moves in order to keep an object or a character in the same region of the image
    \item \textit{Zoom}: The camera operator is zooming in or out
    \item \textit{Pan}: The camera rotates on the horizontal plan
    \item \textit{Tilt}: The camera rotates on the vertical plan
    \item \textit{Dolly}: The camera is being moved using a dolly
    \item \textit{Crane}: Complex camera motion, where both the camera base and the mount are in motion
    \item \textit{Handheld}: The camera operator holds the camera by hand, creating a jerky motion
    \item \textit{Rack focus}: The focus of the lens shifts from one point of the scene to an other
\end{itemize}
Those features are binary for each shot, and a single shot can include different camera motions.

\subsubsection{Camera angle}

In order to convey the emotional states of the characters, or power relationships, filmmakers often use camera angles~\citep{Thomson2009}. For instance, a rolled plan will often indicates that the characters are lost, or in an unstable state of mind, while filming actors with a low angle will give them an impression of power over the other characters, as they tower over the scene. We relied on six different degrees of camera angles~\citep{Wu2017}:
\begin{itemize}
    \item \textit{Eye}: The camera is at the same level as the eyes of the actors
    \item \textit{Low}: The camera is lower than the eyes of the actors, pointing up
    \item \textit{High}: The camera is higher than the eyes of the actors, pointing down
    \item \textit{Worm}: The camera is on the ground, or very low, pointing up with a sharp angle
    \item \textit{Bird}: The camera is very high, pointing down with a sharp angle
    \item \textit{Top}: The camera is at the vertical of the actors, pointing straight down
\end{itemize}

\begin{figure}
    \centering
    \includegraphics[width=\linewidth]{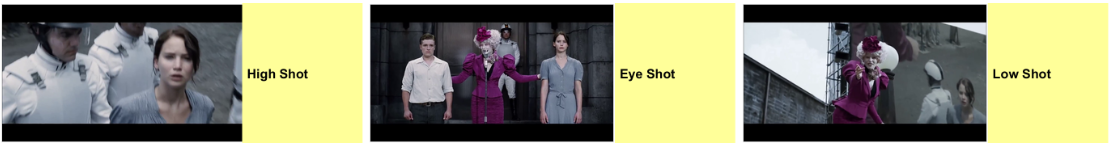}
    \caption{Examples of different camera angles, extracted from \textit{The Hunger Games} (from \cite{Wu2017}).}
    \label{fig:exAngles}
\end{figure}

\subsubsection{Shot size}

\begin{figure}
    \centering
    \includegraphics[width=\linewidth]{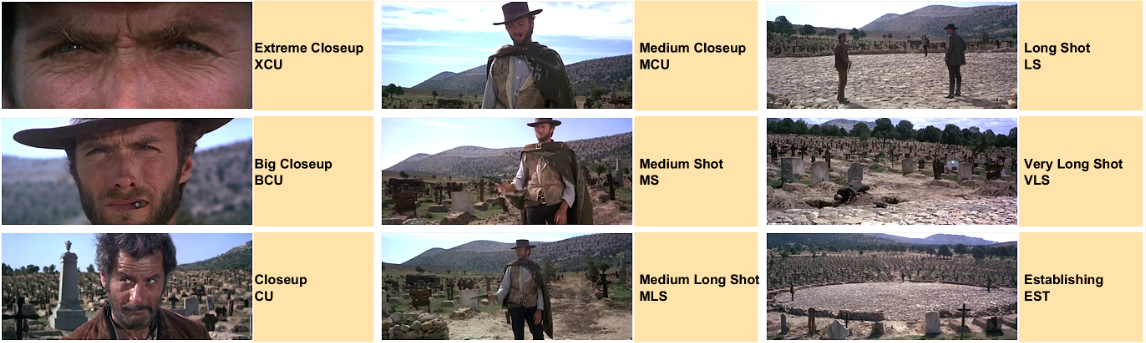}
    \caption{The nine framing sizes, all appearing in \textit{The Good, The Bad \& The Ugly} (from \citet{Wu2017}).}
    \label{fig:exSize}
\end{figure}

The size of a shot represents how close to the camera, for a given lens, the main characters or objects are, and thus how much of their body area is displayed on the screen. Shot size is a way for filmmakers to convey meaning about the importance of a character, for instance, or the tension in a scene. Very large shots can also be used to establish the environment in which the characters will progress. To annotate the shot sizes, we use the 9-size scale defined by~\citet{Thomson2009}. Figure~\ref{fig:exSize} shows the differences between those shot sizes.

\subsubsection{Faces}

As explained by~\citet{Cerf2008}, the presence of faces in images is a very important high-level information to take into account when studying visual attention. We then provide bounding boxes delimiting each face on each frame. Recent state of the art face detection models show that deep learning models extract this information very well. It is then probable that deep visual attention models are also great at extracting faces features, making this hand-crafted feature redundant. However, we include it as it permits an easier analysis of the editing style: for instance, continuity edits will often display faces on the same area of the image, while shot/reverse shots often display faces on opposite sides of the image.

\section{Eye-tracking data collection}
\label{sec:data_collect}

\subsection{Participants and experimental conduct}
We have collected eye-tracking data from 24 volunteers (11 female and 13 male), aged 19 to 56 (average 28.8). Participants were split into two groups, each group watching half of the videos. A few observers were part of both groups, and viewed the whole dataset. In total, we acquired exploitable eye fixation data for 14 participants for each video. Details of the sequences viewed by each group can be found in supplementary material. 

Viewers were ask to fill an explicit consent form, and to perform a pre-test form. The objective of the pre-test form was to detect any kind of visual impairment that could interfere with the conduct of the experiment (colourblindess, or strabism, for instance), as well as ensuring that they could understand English language well enough, as sequences were extracted from the English version of the movies. Participants were informed that they could end the experiment at any moment.

During a session, subjects viewed the 10 movie sequences assigned to their group, in a random order. Sound was delivered by a headset, and volume was set before the first sequence. They could also adjust the volume at will during the experiment. After each sequence, a 15 seconds dark gray screen was displayed. After a series of five clips (around 15 to 20 minutes of video), participants were asked to make a break, as long as they needed, and fill a form, recording whether or not they could recall the scenes they saw, whether or not they had seen the movies previously, or if they recognized any actors in the scenes. After the second series of five clips, at the end of the experiment, they were asked to fill the same form. The total duration of the experiment for a participant was between fifty minutes and one hour.

\subsection{Recording environment and calibration}
Eye movements were recorded using a Tobii X3-120 eye tracker, sampling at 120 Hz. The device was placed at the bottom of a 24,1" screen with a display resolution of $1920 \times 1200$ pixels. All stimuli had the same resolution (96 dpi), and were displayed respecting the original aspect ratio, using letterboxing. The participants were asked to sit at a distance of 65cm from the screen. They were asked to sit as comfortably as possible, in order to minimize head movements. In order to replicate natural viewing conditions, we did not use chin rests. Stereo sound, with a sampling frequency of 44100Hz, was delivered to the participant, using a headset. Calibration was performed using the 9-points Tobii calibration process. In the case of errors of more than one degree, the participant was asked to reposition, and recalibrate. After the break, before viewing the five last clips, participants were asked to validate the previous calibration, and to recalibrate if necessary.

After recording the data for all participants, we used the following cleaning procedure. First, we ensured that every participant had a gaze sampling rate of more than 90\% (i.e. more than 90\% of the sampled points were considered as valid). We then kept only points that were flagged as fixations, eliminating tracking errors due to blinks or other factors, as well as points recorded during saccades. This choice was motivated by the relatively low frequency rate of the eye-tracker, making the analysis of saccadic data impossible. Then, we discarded all points that fell in the letterboxing or outside the screen. Finally, we used the position of the remaining raw points to construct binary fixation maps~: for each frame, we create an image the same size of the frame, where we give the value $1$ to each pixel where a fixation point was flagged during the time the frame was on screen (i.e. $1/24$th of a second), and $0$ to each pixel where no fixation occurred.

\begin{figure}
     \centering
     \begin{minipage}{0.45\linewidth}
         \centering
         \includegraphics[width=\textwidth]{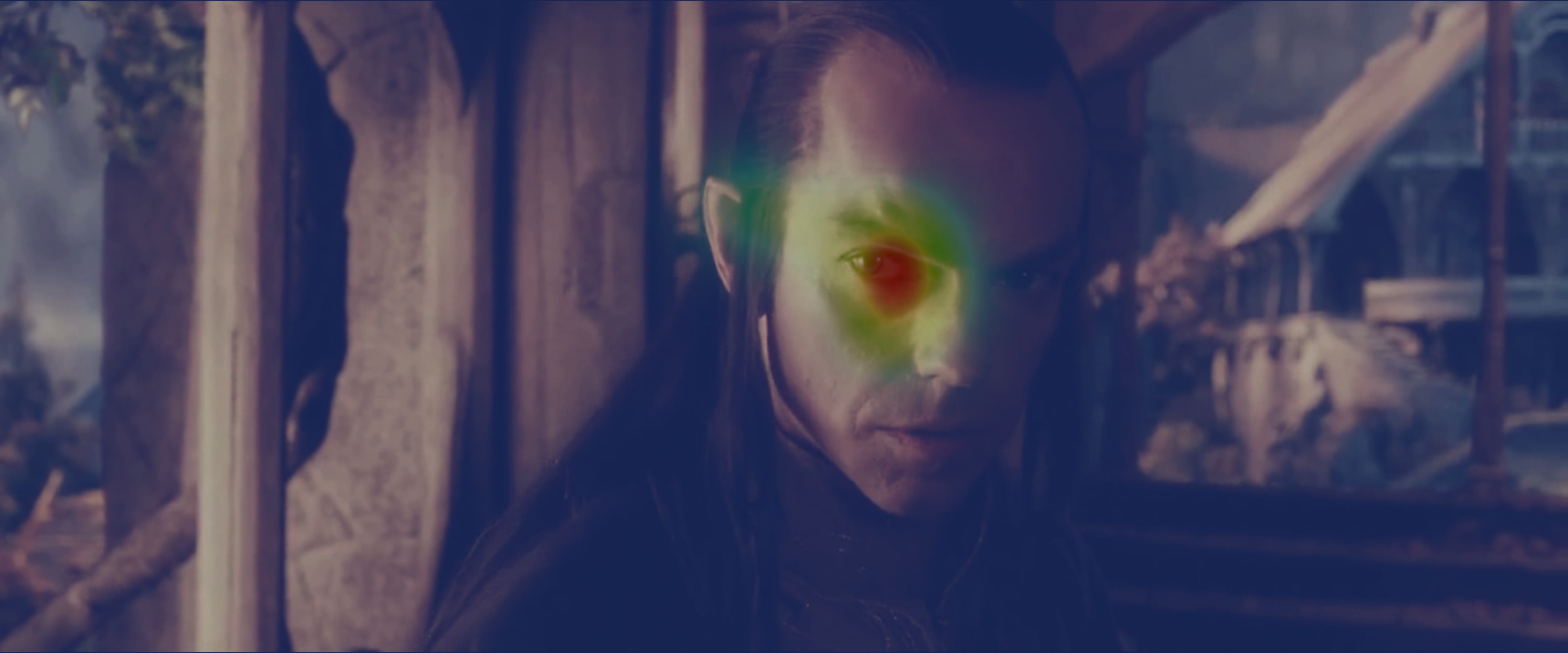}
     \end{minipage}
     \begin{minipage}{0.45\linewidth}
         \centering
         \includegraphics[width=\textwidth]{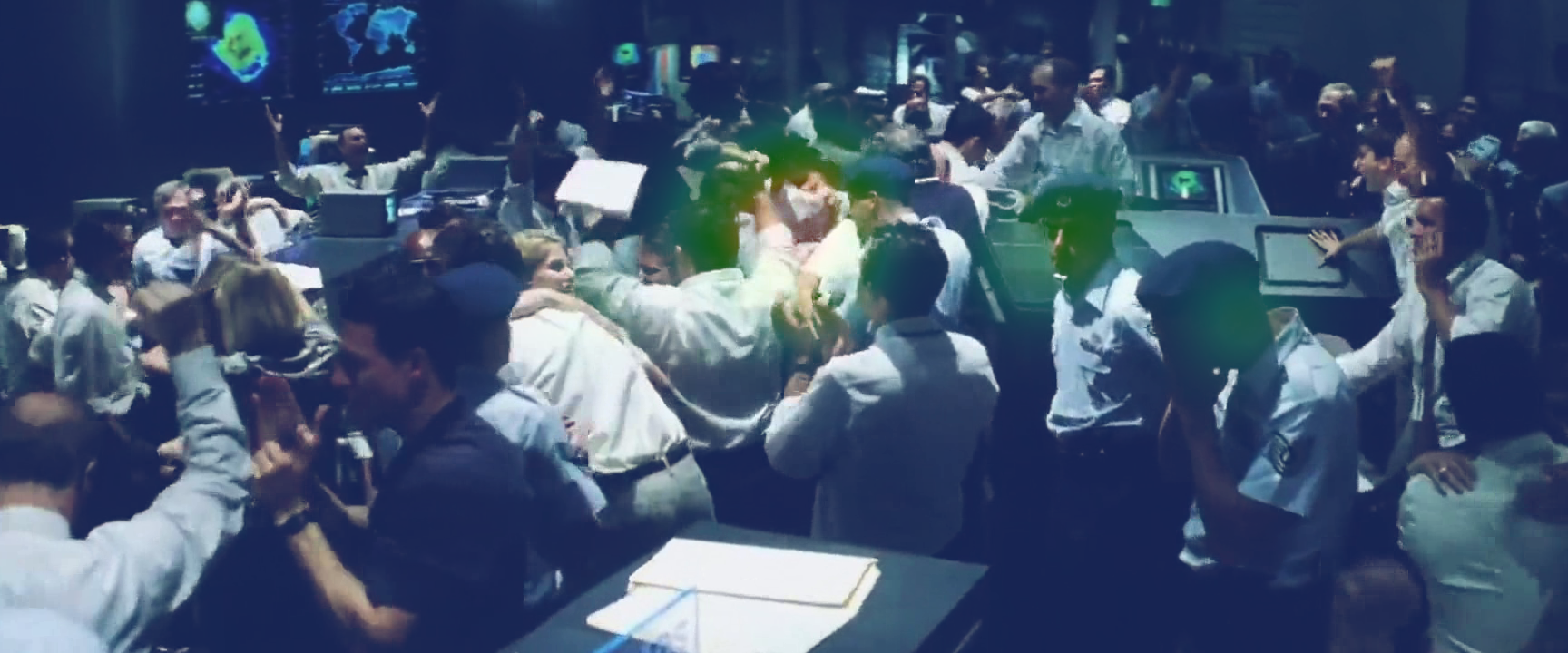}
     \end{minipage}
     \begin{minipage}{0.45\linewidth}
         \centering
         \includegraphics[width=\textwidth]{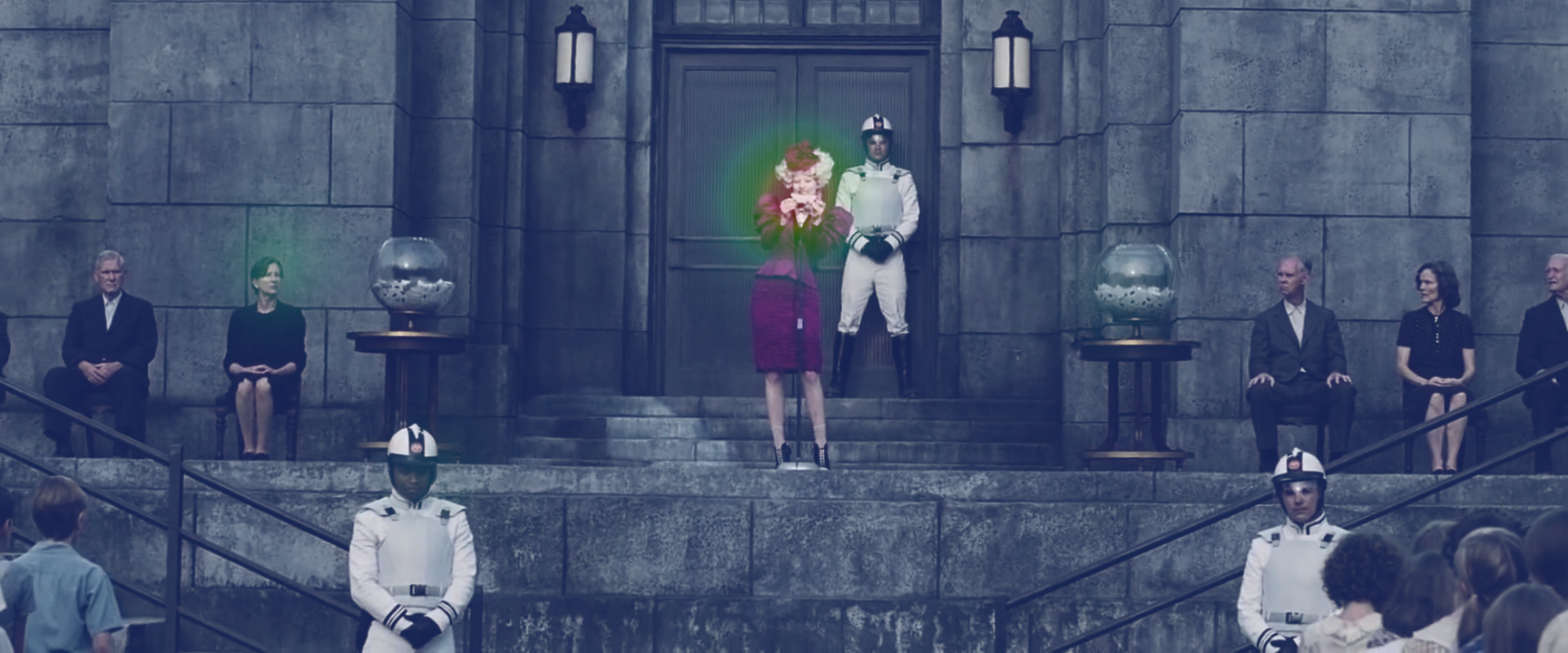}
     \end{minipage}
     \begin{minipage}{0.45\linewidth}
         \centering
         \includegraphics[width=\textwidth]{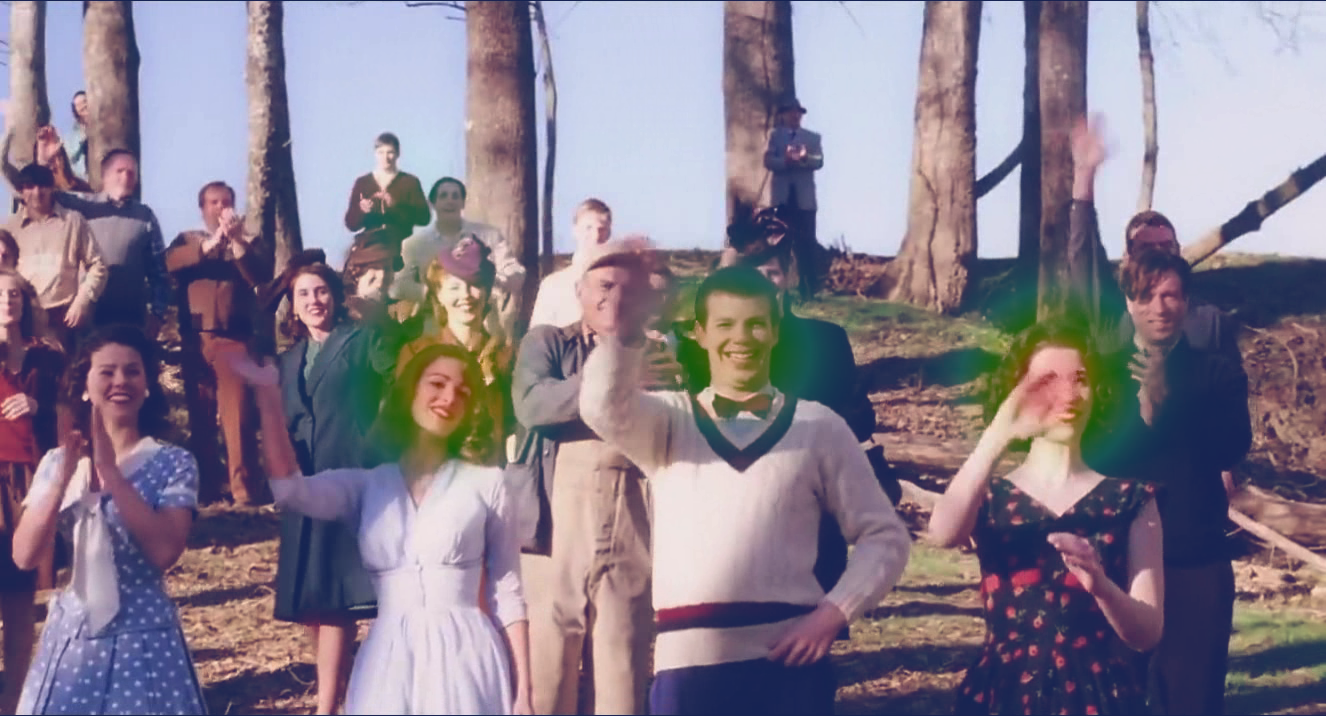}
     \end{minipage}
     \begin{minipage}{0.45\linewidth}
         \centering
         \includegraphics[width=\textwidth]{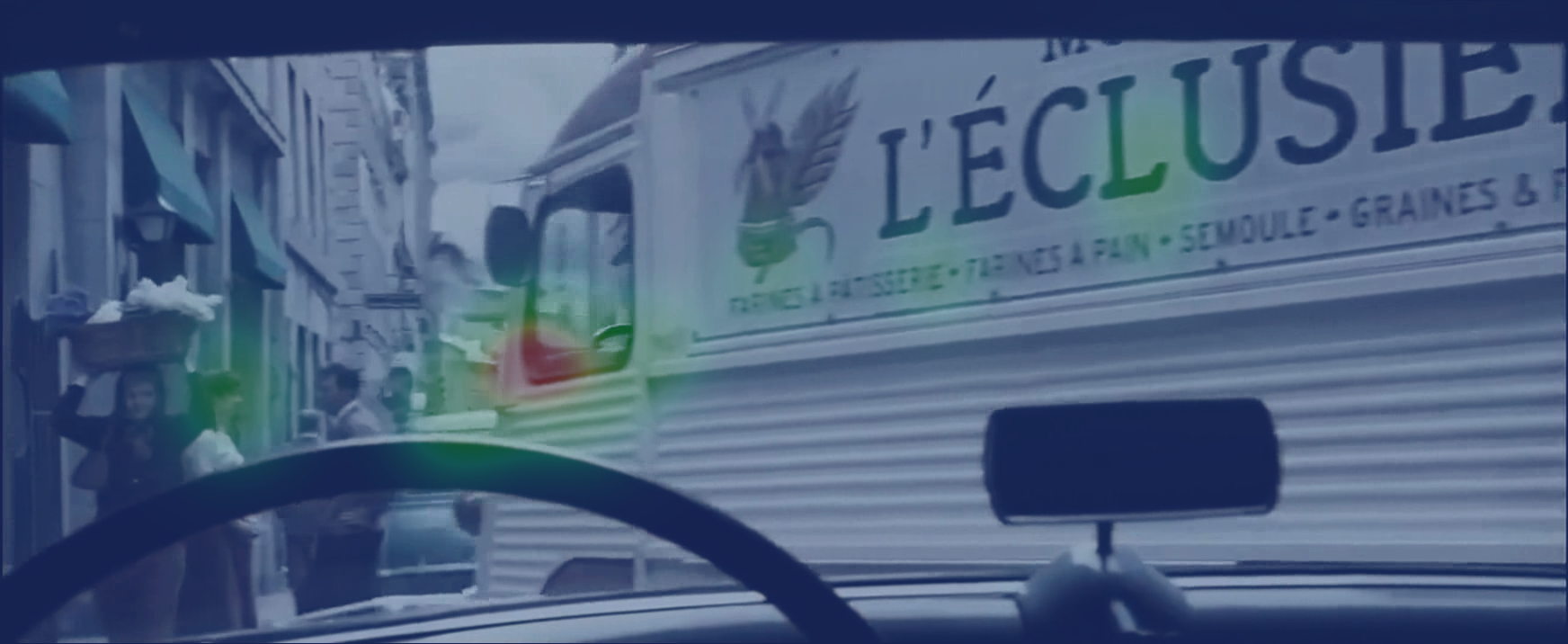}
     \end{minipage}
     \begin{minipage}{0.45\linewidth}
         \centering
         \includegraphics[width=\textwidth]{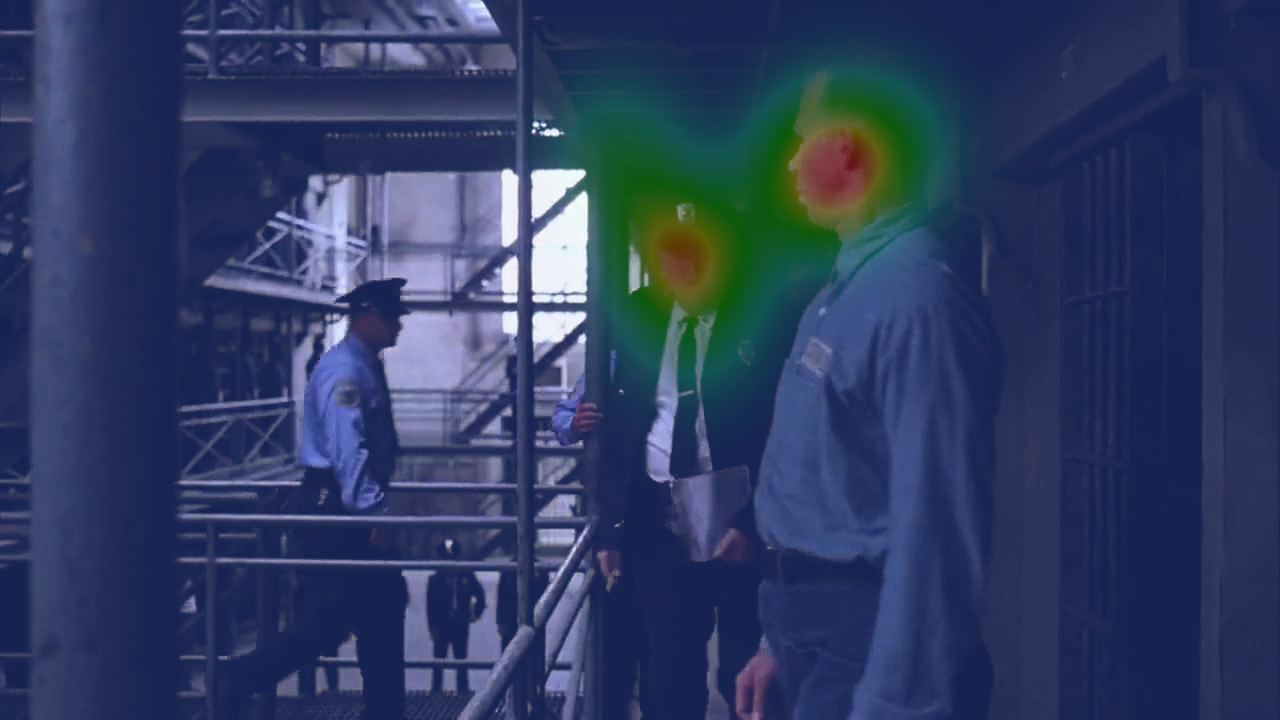}
     \end{minipage}
     \caption{Examples of saliency heatmaps created from the collected fixation points}
     \label{fig:example_saliency}
\end{figure}

\section{Exploring the effects of film making patterns on gaze}
\label{sec:effects_on_gaze}
In this section, we explore several characteristics throughout our database, and analyze underlying relationships between editing patterns and eye fixations patterns. In the following, we will often refer to \textit{fixation maps} and \textit{saliency maps}. For each frame, the fixation map is the binary matrix where each pixel value is $1$ if a fixation occurred at the pixel location during the frame, and $0$ if not, as described previously. Saliency maps are obtained by convolving the fixation maps with a 2-D Gaussian kernel, which variance is set to one degree of visual angle (in our case, one degree of visual angle equals to roughly 45 pixels), in order to approximate the size of the fovea.

\subsection{Editing-induced visual biases}
\begin{figure}[h]
    \centering
    \begin{minipage}{0.3\linewidth}
        \includegraphics[width=\textwidth]{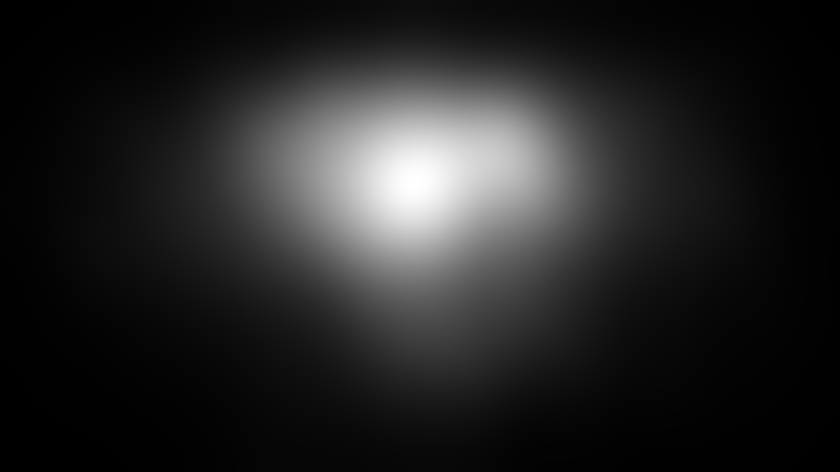}
        \caption*{(a)}
    \end{minipage}
    \quad
    \begin{minipage}{0.3\linewidth}
    \includegraphics[width=\textwidth]{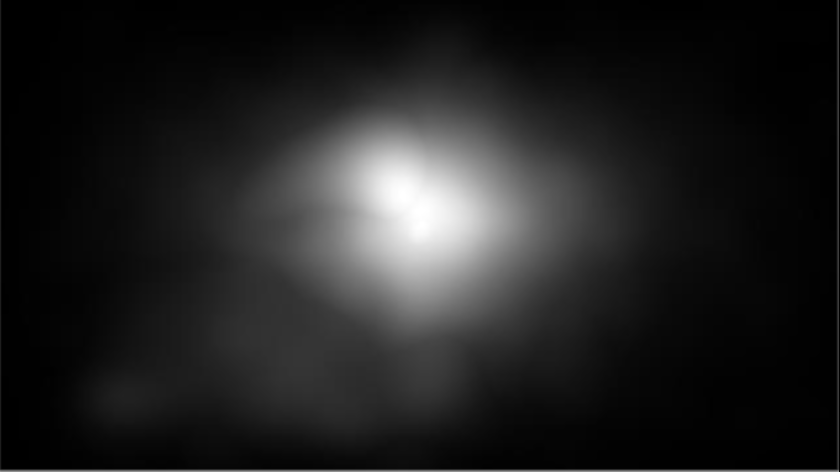}
    \caption*{(b)}
    \end{minipage}
    \quad
    \begin{minipage}{0.3\linewidth}
    \includegraphics[width=\textwidth]{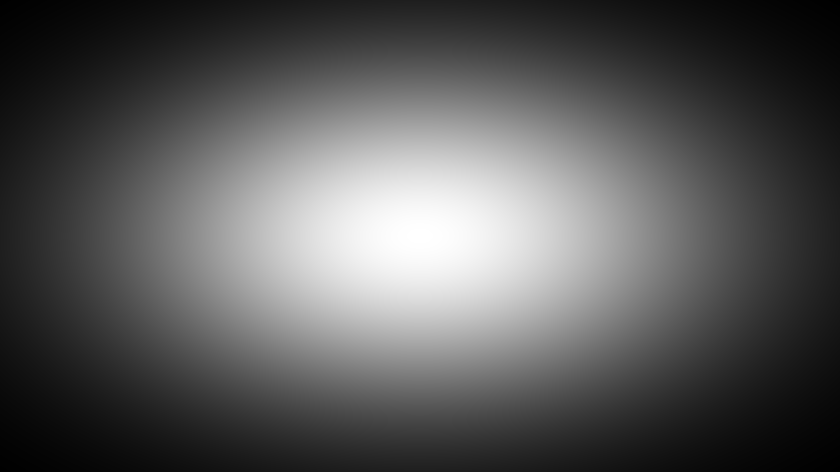}
    \caption*{(c)}
    \end{minipage}
    \caption{Average saliency map of our dataset (a) compared to DHF1K~\citep{Wang2019review} dataset (b) and to a centered gaussian map (c). Both average maps exclude the first 10 frames of each clip.}
    \label{fig:bias_comparaison}
\end{figure}

Studying the average of the saliency maps usually reveals strong attentional biases. For instance, on static images, \citet{tatler2007bias} showed that humans tend to look at the center of the frame. That center bias is also commonly used as a lower baseline for saliency models. In order to avoid recording this bias too much, we did not take into account for our analysis the first 10 frames of each clip, as people tend to look in the middle of the screen before each stimulus. This center bias is also strong on video stimuli: for instance, Fig.~\ref{fig:bias_comparaison} (a) and (b) shows the average saliency map on our dataset and on the DHF1K dataset~\citep{Wang2019review} respectively. However, the latter is composed of Youtube videos, with a great diversity in the content, and no cinematographic scenes, which might cause a different viewing bias. Fig.~\ref{fig:bias_comparaison} (a) shows a peak density slightly above the center of the frame, which would indicate that filmmakers use a different composition rule. Fig.~\ref{fig:bias_comparaison} (c) shows a centered Gaussian map, often used as a baseline for centered bias. Correlation between the average saliency map on our dataset and this centered Gaussian is 0.81, whereas the correlation between the average map on DHF1K and the centered Gaussian is 0.84, which highlights this position discrepancy between the two average saliency maps. This is consistent with the findings of \citet{Breeden2017}, and is most likely due to the rule of thirds~\citep{Brown2016} stating that in cinematography, important elements of the scene should be placed on thirds lines, \emph{i.e.} lines dividing the frame in thirds horizontally and vertically.

\begin{figure}[h]
    \centering
    \begin{minipage}{0.3\linewidth}
        \includegraphics[width=\textwidth]{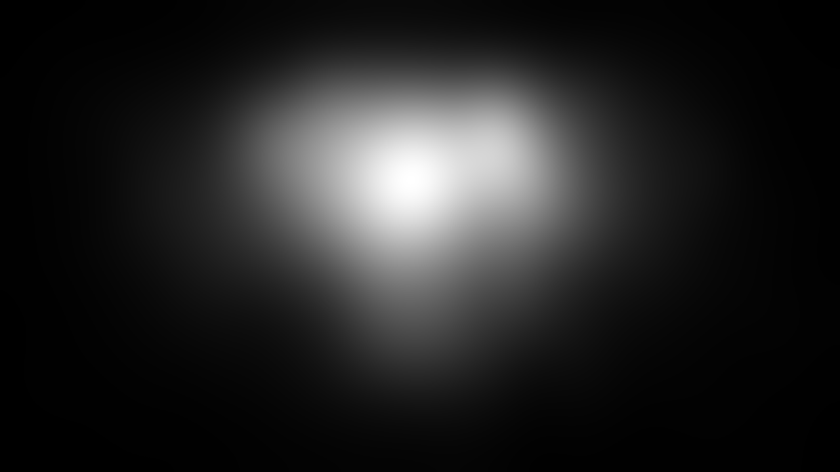}
        \caption*{(a)}
    \end{minipage}
    \quad
    \begin{minipage}{0.3\linewidth}
    \includegraphics[width=\textwidth]{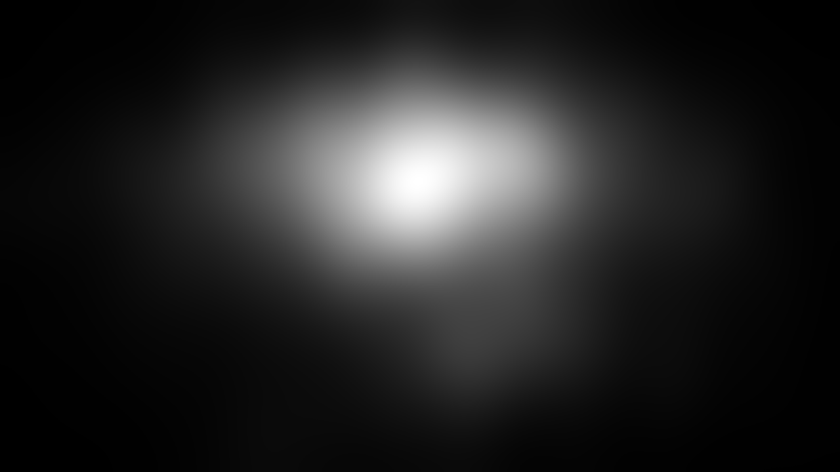}
    \caption*{(b)}
    \end{minipage}
    \quad
    \begin{minipage}{0.3\linewidth}
    \includegraphics[width=\textwidth]{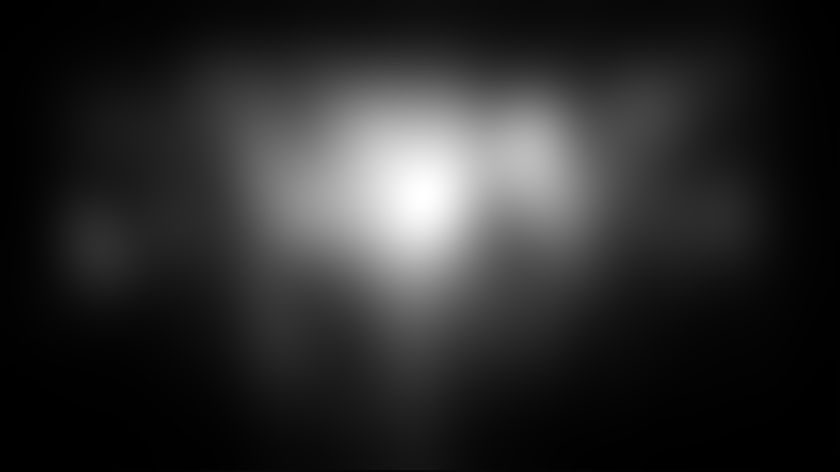}
    \caption*{(c)}
    \end{minipage}
    \quad
    \begin{minipage}{0.3\linewidth}
    \includegraphics[width=\textwidth]{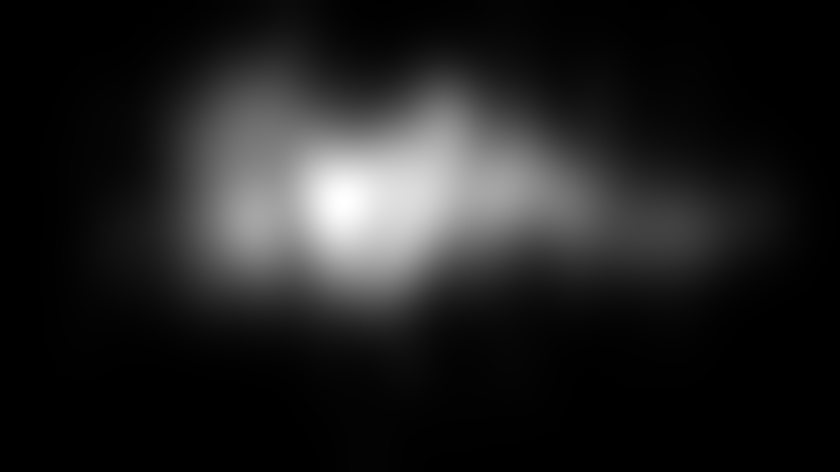}
    \caption*{(d)}
    \end{minipage}
    \quad
    \begin{minipage}{0.3\linewidth}
    \includegraphics[width=\textwidth]{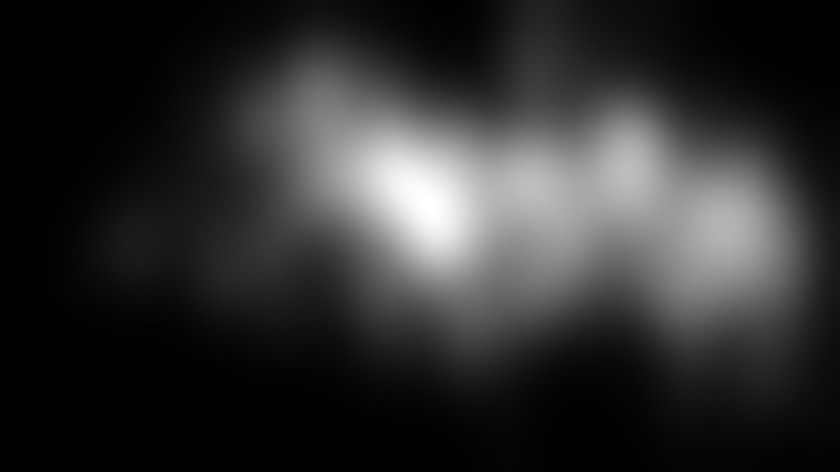}
    \caption*{(e)}
    \end{minipage}
    \caption{Average saliency maps for closeup shots (XCU-BCU-CU) (a), medium shots (MCU-MS-MLS) (b) and long shots (LS-VLS-EST) (b). Subfigure (d) is the average saliency map during pans and dolly shots moving to the left, and (e) is the average saliency map during pans and dolly shots moving to the right.}
    \label{fig:mean_comparaison}
\end{figure}

We also observe disparities in this bias depending on the size of the shot: the wider the shot, the more diffuse that bias is, indicating that directors tend to use a bigger part of the screen area when shooting long shots, while using mostly the center of the frames for important elements during closeups and medium shots (Fig.~\ref{fig:mean_comparaison}, (a,b,c). We also observe a leftward (resp. rightward) bias during pans and dolly shots, where the camera moves towards the left (resp. right), as exposed in Fig.~\ref{fig:mean_comparaison} (d,e). This confirms that camera movements are an important tool for filmmakers to guide the attention of the spectators.

\subsection{Inter-observer visual congruency}
\label{sub_IOC}
Inter-observer congruency (IOC) is a measure of the dispersion of gaze patterns between several observers watching the same stimulus. In other words, it measures how well gaze patterns from a subset of the observers is predictive of the whole set of observers. Thus, it has been used in saliency modeling as an upper baseline. This characteristic is very similar to attentional synchrony~\citep{Smith2008synchrony}, and many methods have been proposed to measure it. For instance, \citet{Goldstein2007} measure how well gaze points are fitted by a single ellipsoid, while \citet{Mital2011} use Gaussian mixture models, associating low cluster covariance to high attentional synchrony. 

In their work, \citet{Breeden2017} use the area of the convex hull of the fixation points, for each frame of their dataset. This allows to take into account all the fixation points, and requires no prior hypothesis about the shape of the regions of interest. However, as they mention, this approach can only be viewed as an upper bound on IOC, as it is very sensitive to outliers. Using it on each frame also does not take into account the temporal aspect of movie viewing: if several observers watch the same two or three points of interest, chances are, if the points are spatially distant from one another, that the convex hull area will be high, even though all the observers exhibited similar gaze patterns in a different order, in terms of fixation locations.

In order to remedy this issue, we used a leave-one-out approach, over a temporal sliding window. Assuming that there is $N$ observers, we gather the locations of all the fixation points of $(N - 1)$ observers during a window of $n$ frames, as well as the locations of the fixation points of the left out observer. We then can build a saliency map using the fixation points of the $N-1$ observers, by convolving the fixation map with a 2-D Gaussian kernel, and use any saliency metric to compare it to the fixation map (or saliency map) of the left out observer. The process is then iterated and averaged over all observers. To compare the saliency map of the $N-1$ observers to the left-out one, we used the Normalized Scanpath Saliency metric (NSS); more details about can be found in \citet{LeMeur2012Metrics}. A high value of this score will mean that people tend to look in the same region, and a low value will indicate a higher dispersion in the fixations. The main drawback of this way of computing IOC, especially for large-scale datasets, is its computational cost, as the process is iterated over every observers, and every $n$-frame window of the dataset.

\begin{figure*}[h]
    \centering
    \includegraphics[width=\textwidth]{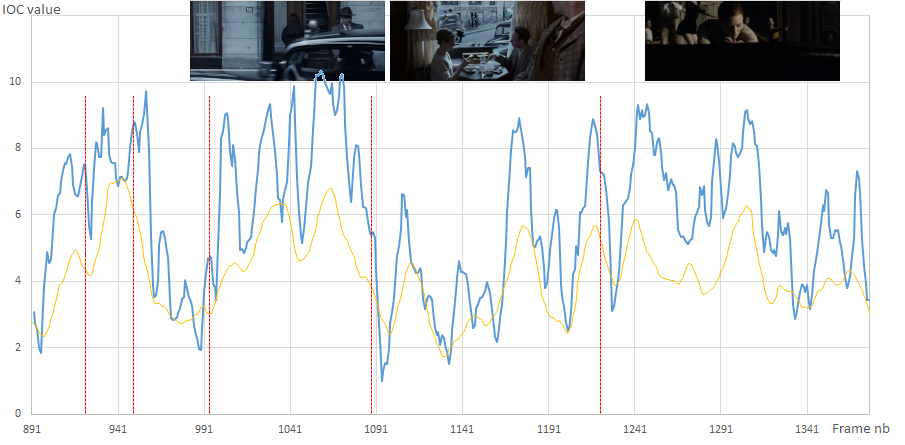}
    \caption{Example of the evolution of inter-observer congruency over a sequence of \textit{Benjamin Button}. Blue is the IOC value computed with a 5-frames time window ($n=5)$, yellow is using a 20-frames time window ($n=20$). Red lines show the locations of the edits; notice the characteristic drop in the IOC value after each cut (for the 5-frames window).}
    \label{fig:IOC_example_ben}
\end{figure*}

The size $n$ of the sliding window can be adjusted, depending on the number of observers, and the studied characteristics. In this work, we chose two window sizes: 5 frames and 20 frames (roughly 200 and 800ms). 5 frames corresponds roughly to the average fixation time, and 20 frames allows for a wider point of view, with less noise. While a shorter window allows for more noise, especially with a relatively short number of observers, it can also underline short-timed patterns. For instance, using the shorter window size, we noticed on every stimulus a significant drop of inter-observer congruency during the five frames consecutive to a cut (see Fig.~\ref{fig:IOC_example_ben}). This would tend to indicate that a short adjustment phase is taking place, as the observers search for the new regions of interest. The amplitude of this drop can be a meaningful perceptual characteristic of a cut, as some editors would voluntarily try to increase that drop, in order to confuse the viewer, or decrease it, in order to create a certain sense of continuity through the cut. Longer window sizes would be better suited for analyzing more general trends, at the scale of a shot, for instance. In the following, we will use the values computed using a 20-frames window, since our editing annotations are at the level of the shot.

\begin{figure}[h]
    \centering
    \includegraphics[width=0.8\linewidth]{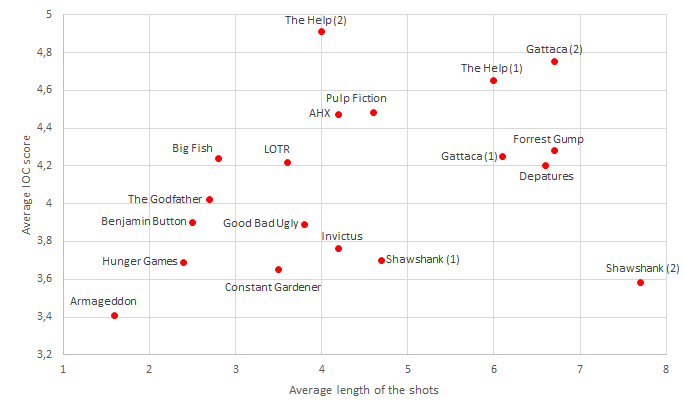}
    \caption{IOC scores depending on the average length of the shots for each sequence (excluding \textit{The Shining}, as it is an outlier in terms of length of the shots)}
    \label{fig:IOC_shot_length}
\end{figure}

As we suspected, the IOC values are relatively high (average of 4.1 over the whole database), especially compared to IOC values in the static case; see for instance \citet{Bruckert2019} for IOC distributions on static datasets. This coroborates the findings of \citet{Goldstein2007} and \citet{Breeden2017}, that viewers only attend to a small portion of the screen area. We also observe a disparity in IOC scores between the movies : the scene with the highest score on average, is the clip from \textit{The Shining} (5.76), and the lowest score on average is the clip from \textit{Armageddon} (3.41). This would tend to indicate that inter-observer congruency reflects certain features in terms of editing style; for instance, Figure~\ref{fig:IOC_shot_length} shows a correlation (0.35) between the average IOC score of a sequence and the average length of the shots of this sequence. However, due to the low number of samples, this correlation is not significant ($p=0.19$).

\subsection{Inter-observer congruency and cinematographic features}

We then considered the effects of directors' choices and editing characteristics on inter-observer congruency. 

\begin{figure}
    \centering
    \begin{minipage}{0.3\linewidth}
        \includegraphics[width=\textwidth]{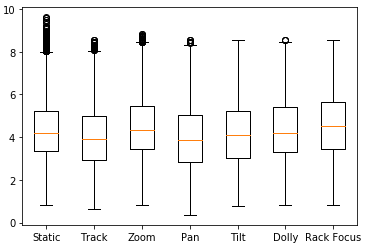}
        \caption*{(a)}
    \end{minipage}
    \quad
    \begin{minipage}{0.3\linewidth}
    \includegraphics[width=\textwidth]{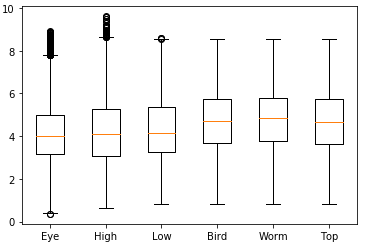}
    \caption*{(b)}
    \end{minipage}
    \quad
    \centering
    \begin{minipage}{0.3\linewidth}
        \includegraphics[width=\textwidth]{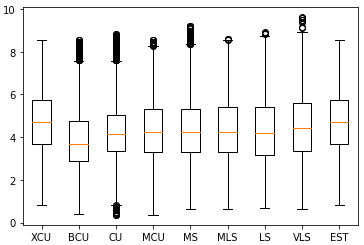}
        \caption*{(c)}
    \end{minipage}
    \quad
    \caption{IOC scores depending on camera movement features (a), camera angles (b) and shot size (c) }
    \label{fig:IOC_comparison}
\end{figure}

Fig.~\ref{fig:IOC_comparison} shows the distributions of IOC scores depending on the high-level annotations described in \ref{annotations}. We performed a one-way ANOVA for each annotation group (camera movements, camera angles and size of the shot), and confirmed that camera movements, angles and shot sizes have a significant influence on IOC scores ($p << 10^{-5}$ in the three cases). Post-hoc pairwise t-tests within the annotation groups show significant differences ($p << 10^{-5}$) between all the characteristics, except between static and dolly shots ($p=0.026$), extreme closeups and establishing shots ($p=0.177$), and all pairs among medium closeups, medium shots, medium-long shots and long shots ($p > 10^{-2}$ in all cases). This might be due to categories sometimes not very well defined, as it can be hard distinguishing between a medium shot and a medium-long shot, for instance. 

\section{Visual attention modeling}
\label{sec:visual_attention_modeling}
In this section, we evaluate several visual saliency models on our database, and highlight certain limitations of current dynamic saliency models. We also discuss how editing patterns can explain some of the failure cases of the models. 

\subsection{Performance results}
In Table~\ref{tab:benchmark_saliency}, we show the performances of state-of-the-art static and dynamic saliency models. In order to evaluate the models, we used the following six classic saliency metrics, described in \citet{LeMeur2012Metrics}:
\begin{itemize}
    \item Pearson's correlation coefficient (CC $\in [-1,1]$) evaluates the degree of linear correlation between the predicted saliency map and the ground truth map.
    \item SIM (SIM $\in [0,1]$) evaluates the similarity between two saliency maps through the intersection between their histograms.
    \item AUC (AUC-J, AUC-B $\in [0,1]$) is the area under the Receiver Operator Curve (ROC). Differences between AUC-J and AUC-B relies on the way true and false positive are computed (see \citet{LeMeur2012Metrics} for more details).
    \item Normalised Scanpath Saliency (NSS $\in [0,+\infty[$) is computed between the predicted saliency map and the ground truth fixation map by measuring the saliency values at the locations of the fixations.
    \item Kullback-Lieber Divergence (KLD $\in [0, +\infty[$) between the two probability distributions represented by the saliency maps.
\end{itemize}

In general, those results are quite low, compared to performances on non-cinematic video datasets (see for instance \citet{Wang2019review}).

\begin{table*}[h]
    \caption{Scores of several saliency models on the database. Non-deep models are marked with *. Best performances are bolded. \dag Note that the testing dataset for the retrained ACLNet model is not exactly the same as the other models, as it is a subset of half of our dataset.}
    \label{tab:benchmark_saliency}
    \centering
    \begin{tabular}{|c|c||c|c|c|c|c|c|}
    \hline
     & Model & CC $\uparrow$ & SIM $\uparrow$ & AUC-J $\uparrow$ & AUC-B $\uparrow$ & NSS $\uparrow$ & KLD $\downarrow$ \\ \hline
     
    Baseline & Center Prior* & 0.398 & 0.302 & 0.859 & 0.771 & 1.762 & 2.490 \\ \hline
    
    \multirow{6}{*}{Dynamic models}
    & PQFT* \citep{Guo2010}  & 0.146 & 0.189 & 0.702 & 0.621 & 0.783 & 2.948 \\ 
    & Two-stream \citep{Bak2018DynamicSaliency} & 0.404 & 0.329 & 0.873 & 0.830 & 1.738 & 1.410 \\ 
    & DeepVS \citep{Jiang2018DeepVS} & 0.457 & 0.361 & 0.880 & 0.829 & 2.270 & 1.245 \\ 
    & ACLNet \citep{Wang2019review} & 0.544 & 0.429 & 0.892 & 0.858 & 2.54 & 1.387 \\ 
    & ACLNet (retrained)\dag & 0.550 & 0.423 & 0.890 & 0.858 & 2.592 & 1.408 \\ 

    & \citet{kao2019model} & \textbf{0.608} & \textbf{0.454} &  \textbf{0.903} & 0.881 & 2.847 & \textbf{1.154}\\ \hline
    \multirow{5}{*}{Static models}
    & Itti* \citep{Itti98} & 0.208 & 0.195 & 0.756 & 0.640 & 1.005 & 2.573 \\
    & SalGAN \citep{Pan2017SalGAN} & 0.533 & 0.390 & 0.897 & 0.781 & 2.622 & 1.372\\
    & DeepGaze II \citep{Kummerer2017DeepGazeII} & 0.584 & 0.362 & 0.846 & 0.774 & \textbf{3.188}  & 2.307\\
    & MSINet \citep{Kroner2020MSINet} & 0.597 & 0.417 & 0.901 & \textbf{0.893} & 2.893 & 1.226 \\\hline
    \end{tabular}
\end{table*}

This would indicate, in the case of deep-learning models, that either the training sets do not contain enough of videos with features specific to cinematic stimuli, or the deep neural networks cannot grasp the information from some of those features. Even though the best performing model is a dynamic one \citep{kao2019model}, we observe that static models (DeepGaze II and MSINet) performances are quite close to those of dynamic models. This might support the latter hypothesis, that dynamic models fail to extract important temporal features.

\begin{figure*}
    \centering
    \includegraphics[width=\textwidth]{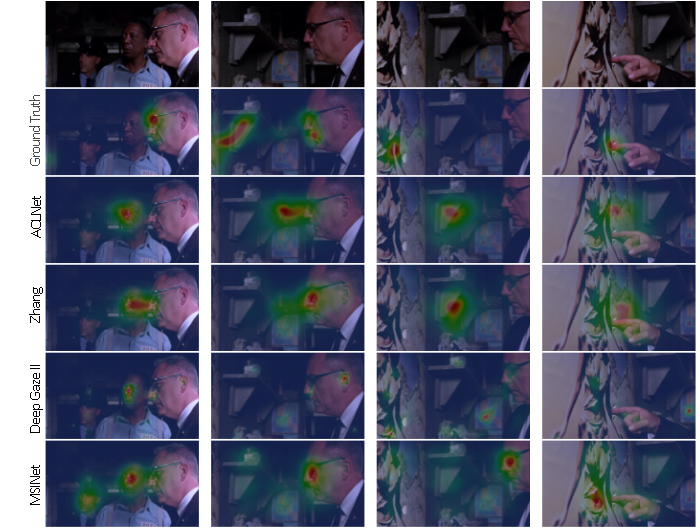}
    \caption{An example of failure case in $Shawshank Redemption$. Here, the camera pans from the face of the prison director to the poster on the wall. While observers quickly shift their attention towards the poster, as suggested by the camera movement, even though it is not yet on screen, models tend to predict areas of interest on the faces.}
    \label{fig:failure_case}
\end{figure*}

Recent work from \citet{Tangemann2020} on the failure cases of saliency models in the context of dynamic stimuli also highlight this point, listing cases like appearing objects, movements or interactions between objects as some of the temporal causes of failure. Figure~\ref{fig:failure_case} shows an example from our database of such a failure case. It should be noted that all the deep learning models are trained on non-cinematic databases, with the exception of ACLNet, which include the Hollywood 2 dataset in its training base. However, this base is not well-fit to learn meaningful cinematographic features, as explained in Section~\ref{sec:datasets}

In order to confirm this hypothesis, we retrained the ACLNet model using the same training procedure described in~\citet{Wang2018Review}. For the static batches, we used the same dataset (SALICON~\citet{Huang2015Salicon}), and for the dynamic batches, we created a set composed of half of our videos, randomly selected, leaving the other videos out for testing (roughly 490000 frames for training, and 450000 frames for test). We only obtained marginally better results on some of the metrics ($0.550$ instead of $0.544$ on the correlation coefficient metric, $2.592$ instead of $2.54$ on the NSS metric), and did not outperform the original model settings on the other. All of this would tend to indicate that some features, specific to cinematographic images, could not be extracted by the model.

\subsection{Edition annotation and model performances}
We also studied how the two best dynamic models, \citet{Zhang2020} and ACLNet \citep{Wang2019review}, performed on our database, depending on shot, camera motion and camera angle characteristics. Table~\ref{tab:annotation_scores} shows the average results of the models depending on the annotation characteristics. Similarly to Subsection~\ref{sub_IOC}, we performed one-way ANOVAs to ensure that results within each table would yield significant differences. In all cases, $p$-values were under $10^{-5}$.
\begin{table*}[h]
    \caption{Scores of two saliency models on the database, depending on hand-crafted editing features. Highest score for each metric and each model is bolded, lowest score is italicized.}
    \label{tab:annotation_scores}
    \centering
    \begin{tabular}{|c|c|c|c|c|c|c|c|c|}
    \hline
     Model & Metric & Static & Track & Zoom & Pan & Tilt & Dolly & Rack Focus \\ \hline
    \multirow{2}{*}{ACLNet}
     &CC  & \textbf{0.561} & 0.545 & 0.538 & \textit{0.466} & 0.488 & 0.517 & 0.545 \\
    &NSS & \textbf{2.631}& 2.610& 2.523 &\textit{2.138} & 2.269& 2.481& 2.610 \\
    \hline
    \multirow{2}{*}{Zhang~\textit{et~al.}}
    & CC& 0.637 & 0.608 & 0.643 & \textit{0.556} & 0.584 & 0.615 & \textbf{0.675} \\
    & NSS & 3.014 & 2.908 & 3.118 & \textit{2.615} & 2.797 & 3.022 & \textbf{3.338}\\\hline
    \end{tabular}
    
    (a) Scores depending on camera motion
    \vspace{5mm}
    
    \centering
    \begin{tabular}{|c|c|c|c|c|c|c|c|}
    \hline
     Model & Metric & Eye & High & Low & Bird & Worm & Top\\ \hline
    \multirow{2}{*}{ACLNet}
     &CC  & 0.552 & \textit{0.500} & 0.525 &\textbf{0.544} &0.532 & 0.540\\
    &NSS & 2.602& \textit{2.343}& 2.465& \textbf{2.699} &2.679 & 2.628\\
    \hline
    \multirow{2}{*}{Zhang~\textit{et~al.}}
    & CC& 0.621 & \textit{0.582} & 0.605 & 0.648 & \textbf{0.679} & 0.672 \\
    & NSS & 2.932 & \textit{2.777} & 2.918 & 3.286 & \textbf{3.513} & 3.375\\\hline
    \end{tabular}
    
    (b) Scores depending on camera angles
    \vspace{5mm}
    
    \centering
    \begin{tabular}{|c|c|c|c|c|c|c|c|c|c|c|}
    \hline
     Model & Metric & XCU & BCU & CU & MCU & MS & MLS & LS & VLS & EST \\ \hline
    \multirow{2}{*}{ACLNet}
     & CC  &0.526 & 0.532 & \textbf{0.586} & 0.549 & 0.497 & 0.510 & \textit{0.473} & 0.520 & 0.512\\
    & NSS & 2.596 & 2.271 & \textbf{2.689} & 2.677 & 2.497 & 2.481 & \textit{2.255} & 2.478 & 2.543 \\
    \hline
    \multirow{2}{*}{Zhang~\textit{et~al.}}
    & CC& 0.656 & 0.607 & \textbf{0.663} & 0.645 & 0.580 & 0.615 & \textit{0.567} & 0.628 & 0.636\\
    & NSS & \textbf{3.320} & \textit{2.679} & 3.099 & 3.186 & 2.889 & 3.027 & 2.733 & 3.089 & 3.221\\\hline
    \end{tabular}
    
    (c) Scores depending on shot size
    \vspace{5mm}
    
\end{table*}

As shown in Table~\ref{tab:annotation_scores} (a), it appears that saliency models perform relatively well on static scenes, or when the camera movement tracks an actor, or an object on screen. Performances are also quite good on shots including rack focuses, which was expected, as this is a very strong tool for the filmmaker to use to direct attention, and deep feature extractors distinguish very well blurry background from clear objects. However, when a more complex camera motion appears, like pans or tilts, models seem to fail more often; this might indicate that saliency models are unable to anticipate that an object is likely to appear in the direction of the motion, which humans usually do. 

With Table~\ref{tab:annotation_scores} (b), we observe that camera angles show little variations in the performances of the models. However, it seems that scenes with high amplitude angles (Bird or Worm) are easier for a model to predict. This is probably due to the fact that those camera angles are often used when filming characters and faces, in order to convey a dominant or a submissive feeling from the characters~\citep{Thomson2009}; since deep learning models are very efficient at recognizing faces, and faces tend to attract gaze, saliency models naturally perform better on those shots.

Finally, looking at Table~\ref{tab:annotation_scores} (c), saliency models seem to exhibit great performances on closeups scenes, which could be, again, because closeup scenes often display faces. Medium to long shots are however harder to predict, maybe because a wider shot allows the director to add more objects or actors on screen, and as shown by \citet{Tangemann2020}, interactions between objects is often a failure case for deep saliency models. Closeup shots also display one of the lowest mean IOC, which could also explain why they are easier to predict.

\section{Conclusion and future work}
\label{sec:conclusion}
In this work, we introduced a new eye-tracking dataset dedicated to study visual attention deployment and eye fixations patterns on cinematographic stimuli. Alongside with the gaze points and saliency data, we provide annotations on several film-specific characteristics, such as camera motion, camera angles or shot size. These annotations allow us to explain a part of the causes of discrepancies between shots in terms of inter-observer visual congruency, and in terms of performances of salien\-cy models.

In particular, we highlight the conclusions of \citet{Tangemann2020} regarding failure cases of state-of-the-art visual attention models. Video stimuli sometimes contain a lot of non-static information, that, in some cases, is more important for directing attention than image-related spatial cues. As directors and editors includes consciously a lot of meaning with their choices of cinematographic parameters (camera motion, choice of the shots within a sequence, shot sizes, etc.), we would advocate  researchers in the field of dynamic saliency to take a closer look at movie sequences, in order to develop different sets of features to explain visual attention.

Looking forward, we can investigate whether or not the high-level cinematic features that we provided would be of help to predict visual deployment, by building a model that includes this kind of metadata at the shot level. Another crucial point that we did not pursue is the context of the shot~: the order of the shots within the sequence has been proven to influence gaze patterns \citep{Loschky2014Jaws,Loschky2020}. As these questions have been tackled from a psychological or cognitive point of view, they remain to be studied by the computer science part of the field, and to be included in visual attention models. This would greatly benefit multiple areas in the image processing field, like video compression for streaming, or automated video description.

Furthermore, we hope that this data would help cinema scholars to quantify potential perceptual reasons to filmmaking conventions, assess continuity editing on sequences and hopefully improve models of automated edition \citep{Galvane2015}.

Finally, developing automated tools to extract similar high-level cinematic information could be particularly of interest, both for the design of such tools, as it would give cues on the way to design better visual attention models on cinematographic data, but also with its outcome, as it would allow the provision of large-scale annotated cinematic databases, which would give a new -- quantitative -- dimension to research on movie contents by cinema scholars.

\section*{Conflict of interest}
The authors declare that they have no conflict of interest.

\section*{Open practices statement}
The study was not preregisered; the dataset generated and analysed during the current study is available at \url{https://github.com/abruckert/eye_tracking_filmmaking}.

\newpage
\bibliographystyle{spbasic} 
\bibliography{biblio.bib}

\begin{thebibliography}{70}
\providecommand{\natexlab}[1]{#1}
\providecommand{\url}[1]{{#1}}
\providecommand{\urlprefix}{URL }
\expandafter\ifx\csname urlstyle\endcsname\relax
  \providecommand{\doi}[1]{DOI~\discretionary{}{}{}#1}\else
  \providecommand{\doi}{DOI~\discretionary{}{}{}\begingroup
  \urlstyle{rm}\Url}\fi
\providecommand{\eprint}[2][]{\url{#2}}

\bibitem[{Bak et~al.(2018)Bak, Kocak, Erdem, and
  Erdem}]{Bak2018DynamicSaliency}
Bak C, Kocak A, Erdem E, Erdem A (2018) Spatio-temporal saliency networks for
  dynamic saliency prediction. IEEE Transactions on Multimedia
  20(7):1688--1698, \doi{10.1109/TMM.2017.2777665}

\bibitem[{Borji(2019)}]{Borji2019Review}
Borji A (2019) Saliency prediction in the deep learning era: Successes and
  limitations. IEEE Transactions on Pattern Analysis and Machine Intelligence
  \doi{10.1109/TPAMI.2019.2935715}

\bibitem[{Borji and Itti(2013)}]{Borji2013Review}
Borji A, Itti L (2013) State-of-the-art in visual attention modeling. IEEE
  Transactions on Pattern Analysis and Machine Intelligence 35(1):185--207,
  \doi{10.1109/TPAMI.2012.89}

\bibitem[{Borji et~al.(2011)Borji, Ahmadabadi, and Araabi}]{Borji2011TopDown}
Borji A, Ahmadabadi MN, Araabi BN (2011) Cost-sensitive learning of top-down
  modulation for attentional control. Machine Vision and Applications
  22:61--76, \doi{10.1007/s00138-009-0192-0}

\bibitem[{Borji et~al.(2013)Borji, Sihite, and Itti}]{Borji2013}
Borji A, Sihite DN, Itti L (2013) Quantitative analysis of human-model
  agreement in visual saliency modeling: A comparative study. IEEE Transactions
  on Image Processing 22(1):55--69, \doi{10.1109/TIP.2012.2210727}

\bibitem[{Breathnach(2016)}]{Breathnach2016AttentionalSA}
Breathnach D (2016) Attentional synchrony and the affects of repetitve movie
  viewing. In: AICS,
  \urlprefix\url{http://ceur-ws.org/Vol-1751/AICS_2016_paper_57.pdf}

\bibitem[{Breeden and Hanrahan(2017)}]{Breeden2017}
Breeden K, Hanrahan P (2017) Gaze data for the analysis of attention in feature
  films. ACM Transactions on Applied Perception 14(4), \doi{10.1145/3127588}

\bibitem[{Brown(2016)}]{Brown2016}
Brown B (2016) Cinematography: theory and practice: image making for
  cinematographers and directors

\bibitem[{Bruce and Tsotsos(2005)}]{Bruce2005}
Bruce NDB, Tsotsos JK (2005) Saliency based on information maximization. In:
  Proceedings of the 18th International Conference on Neural Information
  Processing Systems, NIPS'05, pp 155--–162, \doi{10.5555/2976248.2976268}

\bibitem[{Bruckert et~al.(2019)Bruckert, Lam, Christie, and
  {Le~Meur}}]{Bruckert2019}
Bruckert A, Lam YH, Christie M, {Le~Meur} O (2019) Deep learning for
  inter-observer congruency prediction. In: 2019 IEEE International Conference
  on Image Processing (ICIP), pp 3766--3770, \doi{10.1109/ICIP.2019.8803596}

\bibitem[{Cerf et~al.(2008)Cerf, Harel, Einhaeuser, and Koch}]{Cerf2008}
Cerf M, Harel J, Einhaeuser W, Koch C (2008) Predicting human gaze using
  low-level saliency combined with face detection. In: Advances in Neural
  Information Processing Systems, vol~20, pp 241--248

\bibitem[{Chua et~al.(2005)Chua, Boland, and Nisbett}]{Chua2005}
Chua HF, Boland JE, Nisbett RE (2005) Cultural variation in eye movements
  during scene perception. Proceedings of the National Academy of Sciences
  102(35):12629--12633, \doi{10.1073/pnas.0506162102}

\bibitem[{Cornia et~al.(2018)Cornia, Baraldi, Serra, and
  Cucchiara}]{Cornia2018SAM}
Cornia M, Baraldi L, Serra G, Cucchiara R (2018) Predicting human eye fixations
  via an lstm-based saliency attentive model. IEEE Transactions on Image
  Processing 27(10):5142--5154, \doi{10.1109/TIP.2018.2851672}

\bibitem[{Cutting and Armstrong(2016)}]{Cutting2016}
Cutting JE, Armstrong KL (2016) Facial expression, size, and clutter:
  Inferences from movie structure to emotion judgments and back. Attention,
  Perception, \& Psychophysics 78:891–--901, \doi{10.3758/s13414-015-1003-5}

\bibitem[{Duchowski(2002)}]{Duchowski2002}
Duchowski AT (2002) A breadth-first survey of eye-tracking applications.
  Behavior Research Methods, Instruments, \& Computers 34:455--470,
  \doi{10.3758/BF03195475}

\bibitem[{Findlay(1997)}]{Findlay1997}
Findlay JM (1997) Saccade target selection during visual search. Vision
  Research 37(5):617--631, \doi{10.1016/S0042-6989(96)00218-0}

\bibitem[{Foulsham and Underwood(2008)}]{Foulsham2008}
Foulsham T, Underwood G (2008) What can saliency models predict about eye
  movements? spatial and sequential aspects of fixations during encoding and
  recognition. Journal of Vision 8(2), \doi{10.1167/8.2.6}

\bibitem[{Galvane et~al.(2015)Galvane, Christie, Lino, and
  Ronfard}]{Galvane2015}
Galvane Q, Christie M, Lino C, Ronfard R (2015) Camera-on-rails: Automated
  computation of constrained camera paths. In: Proceedings of the 8th ACM
  SIGGRAPH Conference on Motion in Games, pp 151–--157,
  \doi{10.1145/2822013.2822025}

\bibitem[{Gao et~al.(2009)Gao, Han, and Vasconcelos}]{Gao2009}
Gao D, Han S, Vasconcelos N (2009) Discriminant saliency, the detection of
  suspicious coincidences, and applications to visual recognition. IEEE
  Transactions on Pattern Analysis and Machine Intelligence 31(6):989--1005,
  \doi{10.1109/TPAMI.2009.27}

\bibitem[{Gitman et~al.(2014)Gitman, Erofeev, Vatolin, Bolshakov, and
  Fedorov}]{Gitman2014Savam}
Gitman Y, Erofeev M, Vatolin D, Bolshakov A, Fedorov A (2014) Semiautomatic
  {Visual-Attention} modeling and its application to video compression. In:
  2014 IEEE International Conference on Image Processing (ICIP), pp 1105--1109,
  \doi{10.1109/ICIP.2014.7025220}

\bibitem[{Goldstein et~al.(2007)Goldstein, Woods, and Peli}]{Goldstein2007}
Goldstein RB, Woods RL, Peli E (2007) Where people look when watching movies:
  do all viewers look at the same place? Computers in Biology and Medicine
  37(7):957–--964, \doi{10.1016/j.compbiomed.2006.08.018}

\bibitem[{Gorji and Clark(2018)}]{Gorji2018}
Gorji S, Clark JJ (2018) Going from image to video saliency: Augmenting image
  salience with dynamic attentional push. In: 2018 IEEE Conference on Computer
  Vision and Pattern Recognition (CVPR), pp 7501--7511,
  \doi{10.1109/CVPR.2018.00783}

\bibitem[{Guo and Zhang(2010)}]{Guo2010}
Guo C, Zhang L (2010) A novel multiresolution spatiotemporal saliency detection
  model and its applications in image and video compression. IEEE Transactions
  on Image Processing 19(1):185--198, \doi{10.1109/TIP.2009.2030969}

\bibitem[{Hadizadeh and Bajic(2014)}]{Hadizadeh2014}
Hadizadeh H, Bajic IV (2014) Saliency-aware video compression. IEEE
  Transactions on Image Processing 23:19--33, \doi{10.1109/TIP.2013.2282897}

\bibitem[{Harel et~al.(2006)Harel, Koch, and Perona}]{Harel2006GBVS}
Harel J, Koch C, Perona P (2006) Graph-based visual saliency. In: Proceedings
  of the 19th International Conference on Neural Information Processing
  Systems, NIPS'06, pp 545–--552, \doi{10.5555/2976456.2976525}

\bibitem[{Harezlak and Kasprowski(2018)}]{Harezlak2018}
Harezlak K, Kasprowski P (2018) Application of eye tracking in medicine: A
  survey, research issues and challenges. In: Computerized Medical Imaging and
  Graphics, vol~65, pp 176--190, \doi{10.1016/j.compmedimag.2017.04.006}

\bibitem[{Harezlak et~al.(2016)Harezlak, Kasprowski, Dzierzega, and
  Kruk}]{Harezlak2016}
Harezlak K, Kasprowski P, Dzierzega M, Kruk K (2016) Application of eye
  tracking for diagnosis and therapy of children with brain disabilities. In:
  Intelligent Decision Technologies, pp 323--333,
  \doi{10.1007/978-3-319-39627-9_28}

\bibitem[{Huang et~al.(2015)Huang, Shen, Boix, and Zhao}]{Huang2015Salicon}
Huang X, Shen C, Boix X, Zhao Q (2015) Salicon: Reducing the semantic gap in
  saliency prediction by adapting deep neural networks. In: 2015 IEEE
  International Conference on Computer Vision (ICCV), pp 262--270,
  \doi{10.1109/ICCV.2015.38}

\bibitem[{Itti et~al.(1998)Itti, Koch, and Niebur}]{Itti98}
Itti L, Koch C, Niebur E (1998) A model of saliency-based visual attention for
  rapid scene analysis. IEEE Transactions on Pattern Analysis and Machine
  Intelligence 20(11):1254--1259, \doi{10.1109/34.730558}

\bibitem[{Jiang et~al.(2018)Jiang, Xu, Liu, Qiao, and Wang}]{Jiang2018DeepVS}
Jiang L, Xu M, Liu T, Qiao M, Wang Z (2018) Deepvs: A deep learning based video
  saliency prediction approach. In: 2018 European Conference on Computer Vision
  (ECCV), pp 625--642, \doi{10.1007/978-3-030-01264-9_37}

\bibitem[{Jodogne and Piater(2007)}]{Jodogne2007TopDown}
Jodogne S, Piater JH (2007) Closed-loop learning of visual control policies.
  Journal of Artificial Intelligence Research 28(1):349–--391,
  \doi{10.5555/1622591.1622601}

\bibitem[{Kanan et~al.(2009)Kanan, Tong, Zhang, and Cottrell}]{Kanan2009}
Kanan C, Tong MH, Zhang L, Cottrell GW (2009) Sun: Top-down saliency using
  natural statistics. Visual cognition 17(6-7):979--–1003,
  \doi{10.1080/13506280902771138}

\bibitem[{Karessli et~al.(2017)Karessli, Akata, Schiele, and
  Bulling}]{Karessli2017}
Karessli N, Akata Z, Schiele B, Bulling A (2017) Gaze embeddings for zero-shot
  image classification. In: Proceedings of the IEEE Conference on Computer
  Vision and Pattern Recognition (CVPR), pp 4525--4534,
  \doi{10.1109/CVPR.2017.679}

\bibitem[{Koehler et~al.(2014)Koehler, Guo, Zhang, and Eckstein}]{Koehler2014}
Koehler K, Guo F, Zhang S, Eckstein MP (2014) What do saliency models predict?
  Journal of Vision 14(3), \doi{10.1167/14.3.14}

\bibitem[{Kroner et~al.(2020)Kroner, Senden, Driessens, and
  Goebel}]{Kroner2020MSINet}
Kroner A, Senden M, Driessens K, Goebel R (2020) Contextual encoder–decoder
  network for visual saliency prediction. Neural Networks 129:261--270,
  \doi{10.1016/j.neunet.2020.05.004}

\bibitem[{Kruthiventi et~al.(2017)Kruthiventi, Ayush, and
  Babu}]{Kruthiventi2015DeepFix}
Kruthiventi SSS, Ayush K, Babu RV (2017) Deepfix: A fully convolutional neural
  network for predicting human eye fixations. IEEE Transactions on Image
  Processing 26(9):4446--4456, \doi{10.1109/TIP.2017.2710620}

\bibitem[{K{\"u}mmerer et~al.(2017)K{\"u}mmerer, Wallis, Gatys, and
  Bethge}]{Kummerer2017DeepGazeII}
K{\"u}mmerer M, Wallis TSA, Gatys LA, Bethge M (2017) Understanding low- and
  high-level contributions to fixation prediction. In: 2017 IEEE International
  Conference on Computer Vision (ICCV), pp 4789--4798,
  \doi{10.1109/ICCV.2017.513}

\bibitem[{{Le~Meur} and Baccino(2013)}]{LeMeur2012Metrics}
{Le~Meur} O, Baccino T (2013) Methods for comparing scanpaths and saliency
  maps: strengths and weaknesses. Behavior Research Methods 45(1):251--266,
  \doi{10.3758/s13428-012-0226-9}

\bibitem[{{Le~Meur} et~al.(2006){Le~Meur}, {Le~Callet}, Barba, and
  Thoreau}]{LeMeur2006}
{Le~Meur} O, {Le~Callet} P, Barba D, Thoreau D (2006) A coherent computational
  approach to model bottom-up visual attention. IEEE Transactions on Pattern
  Analysis and Machine Intelligence 28(5):802--817, \doi{10.1109/TPAMI.2006.86}

\bibitem[{{Le Meur} et~al.(2017){Le Meur}, Coutrot, Liu, R{\"a}m{\"a}, {Le
  Roch}, and Helo}]{LeMeur2017}
{Le Meur} O, Coutrot A, Liu Z, R{\"a}m{\"a} P, {Le Roch} A, Helo A (2017)
  Visual attention saccadic models learn to emulate gaze patterns from
  childhood to adulthood. IEEE Transactions on Image Processing
  26(10):4777--4789, \doi{10.1109/TIP.2017.2722238}

\bibitem[{Loschky et~al.(2014)Loschky, Larson, Magliano, and
  Smith}]{Loschky2014Jaws}
Loschky L, Larson A, Magliano J, Smith TJ (2014) What would jaws do? the
  tyranny of film and the relationship between gaze and higher-level
  comprehension processes for narrative film. Journal of Vision 14(10),
  \doi{10.1167/14.10.761}

\bibitem[{Loschky et~al.(2020)Loschky, Larson, Smith, and
  Magliano}]{Loschky2020}
Loschky LC, Larson AM, Smith TJ, Magliano JP (2020) The scene perception \&
  event comprehension theory (spect) applied to visual narratives. Topics in
  Cognitive Science 12(1):311--351, \doi{10.1111/tops.12455}

\bibitem[{Mahadevan and Vasconcelos(2010)}]{Mahadevan2010}
Mahadevan V, Vasconcelos N (2010) Spatiotemporal saliency in dynamic scenes.
  IEEE Transactions on Pattern Analysis and Machine Intelligence
  32(1):171--177, \doi{10.1109/TPAMI.2009.112}

\bibitem[{Majaranta and R{\"a}ih{\"a}(2002)}]{Majaranta2002}
Majaranta P, R{\"a}ih{\"a} KJ (2002) Twenty years of eye typing: Systems and
  design issues. In: Proceedings of the 2002 Symposium on Eye Tracking Research
  \& Applications, ETRA '02, pp 15–--22, \doi{10.1145/507072.507076}

\bibitem[{Mathe and Sminchisescu(2015)}]{Mathe2015Hollywood2}
Mathe S, Sminchisescu C (2015) Actions in the eye: Dynamic gaze datasets and
  learnt saliency models for visual recognition. IEEE Transactions on Pattern
  Analysis and Machine Intelligence 37(7):1408--1424,
  \doi{10.1109/TPAMI.2014.2366154}

\bibitem[{Mital et~al.(2011)Mital, Smith, Hill, and Henderson}]{Mital2011}
Mital PK, Smith TJ, Hill RL, Henderson JM (2011) Clustering of gaze during
  dynamic scene viewing is predicted by motion. Cognitive Computation
  3(1):5--–24, \doi{10.1007/s12559-010-9074-z}

\bibitem[{Pan et~al.(2016)Pan, Sayrol, Giro-I-Nieto, McGuinness, and
  O’Connor}]{Pan2016DeepNet}
Pan J, Sayrol E, Giro-I-Nieto X, McGuinness K, O’Connor NE (2016) Shallow and
  deep convolutional networks for saliency prediction. In: 2016 IEEE Conference
  on Computer Vision and Pattern Recognition (CVPR), pp 598--606,
  \doi{10.1109/CVPR.2016.71}

\bibitem[{Pan et~al.(2017)Pan, Canton, McGuinness, O'Connor, Torres, Sayrol,
  and i~Nieto}]{Pan2017SalGAN}
Pan J, Canton C, McGuinness K, O'Connor NE, Torres J, Sayrol E, i~Nieto XG
  (2017) Salgan: Visual saliency prediction with generative adversarial
  networks. \urlprefix\url{https://arxiv.org/abs/1701.01081}

\bibitem[{Rahman et~al.(2014)Rahman, Pellerin, and Houzet}]{Rahman2014}
Rahman A, Pellerin D, Houzet D (2014) Influence of number, location and size of
  faces on gaze in video. Journal of Eye Movement Research 7(2):891–--901,
  \doi{10.16910/jemr.7.2.5}

\bibitem[{Rayner et~al.(2009)Rayner, Castelhano, and Yang}]{Rayner2009}
Rayner K, Castelhano MS, Yang J (2009) Eye movements when looking at
  unusual/weird scenes: Are there cultural differences? Journal of Experimental
  Psychology: Learning, Memory, and Cognition 35(1):254--259,
  \doi{10.1037/a0013508}

\bibitem[{Ronfard et~al.(2013)Ronfard, Gandhi, and Boiron}]{Ronfard2013}
Ronfard R, Gandhi V, Boiron L (2013) The prose storyboard language : A tool for
  annotating and directing movies. AAAI Workshop on Intelligent Cinematography
  and Editing

\bibitem[{Rudoy et~al.(2013)Rudoy, Goldman, Shechtman, and
  Zelnik-Manor}]{Rudoy2013}
Rudoy D, Goldman DB, Shechtman E, Zelnik-Manor L (2013) Learning video saliency
  from human gaze using candidate selection. In: 2013 IEEE Conference on
  Computer Vision and Pattern Recognition (CVPR), pp 1147--1154,
  \doi{10.1109/CVPR.2013.152}

\bibitem[{Smith(2013)}]{Smith2013Review}
Smith TJ (2013) Watchin you watch movies : Using eye tracking to inform
  cognitive film theory, pp 165--192.
  \doi{10.1093/acprof:oso/9780199862139.003.0009}

\bibitem[{Smith and Henderson(2008)}]{Smith2008synchrony}
Smith TJ, Henderson J (2008) Attentional synchrony in static and dynamic
  scenes. Journal of Vision 8(6):774, \doi{10.1167/8.6.773}

\bibitem[{Smith and Mital(2013)}]{Smith2013Synchrony}
Smith TJ, Mital PK (2013) Attentional synchrony and the influence of viewing
  task on gaze behavior in static and dynamic scenes. Journal of Vision
  13(8):16--16, \doi{10.1167/13.8.16}

\bibitem[{Smith et~al.(2012)Smith, Levin, and Cutting}]{Smith2012Review}
Smith TJ, Levin D, Cutting JE (2012) A window on reality: Perceiving edited
  moving images. Current Directions in Psychological Science 21(2):107--113,
  \doi{10.1177/0963721412437407}

\bibitem[{Tangemann et~al.(2020)Tangemann, K{\"u}mmerer, Wallis, and
  Bethge}]{Tangemann2020}
Tangemann M, K{\"u}mmerer M, Wallis TSA, Bethge M (2020) Measuring the
  importance of temporal features in video saliency. In: 2020 European
  Conference on Computer Vision (ECCV), pp 667--684,
  \doi{10.1007/978-3-030-58604-1_40}

\bibitem[{Tatler(2007)}]{tatler2007bias}
Tatler BW (2007) The central fixation bias in scene viewing: Selecting an
  optimal viewing position independently of motor biases and image feature
  distributions. Journal of Vision 7(14):4--4, \doi{10.1167/7.14.4}

\bibitem[{Tavakoli et~al.(2019)Tavakoli, Borji, Rahtu, and
  Kannala}]{Tavakoli2019DAVE}
Tavakoli HR, Borji A, Rahtu E, Kannala J (2019) Dave: A deep audio-visual
  embedding for dynamic saliency prediction.
  \urlprefix\url{https://arxiv.org/abs/1905.10693}

\bibitem[{Thompson and Bowen(2009)}]{Thomson2009}
Thompson R, Bowen C (2009) Grammar of the Shot

\bibitem[{Valuch and Ansorge(2015)}]{Valuch2015}
Valuch C, Ansorge U (2015) The influence of color during continuity cuts in
  edited movies: an eye-tracking study. Multimedia Tools and Applications
  74:10161--10176, \doi{10.1007/s11042-015-2806-z}

\bibitem[{Vig et~al.(2014)Vig, Dorr, and Cox}]{Vig2014}
Vig E, Dorr M, Cox D (2014) Large-scale optimization of hierarchical features
  for saliency prediction in natural images. In: 2014 IEEE Conference on
  Computer Vision and Pattern Recognition (CVPR), pp 2798--2805,
  \doi{10.1109/CVPR.2014.358}

\bibitem[{Wang et~al.(2018)Wang, Shen, Guo, Cheng, and Borji}]{Wang2018Review}
Wang W, Shen J, Guo F, Cheng MM, Borji A (2018) Revisiting video saliency: A
  large-scale benchmark and a new model. In: Proceedings of the IEEE Conference
  on Computer Vision and Pattern Recognition (CVPR), pp 4894--4903,
  \doi{10.1109/CVPR.2018.00514}

\bibitem[{Wang et~al.(2019)Wang, Shen, Xie, Cheng, Ling, and
  Borji}]{Wang2019review}
Wang W, Shen J, Xie J, Cheng MM, Ling H, Borji A (2019) Revisiting video
  saliency prediction in the deep learning era. IEEE Transactions on Pattern
  Analysis and Machine Intelligence \doi{10.1109/TPAMI.2019.2924417}

\bibitem[{Wu et~al.(2017)Wu, Galvane, Lino, and Christie}]{Wu2017}
Wu HY, Galvane Q, Lino C, Christie M (2017) Analyzing elements of style in
  annotated film clips. In: Eurographics Workshop on Intelligent Cinematography
  and Editing, \doi{10.2312/wiced.20171068}

\bibitem[{Wu et~al.(2018)Wu, Pal\`{u}, Ranon, and Christie}]{Wu2018}
Wu HY, Pal\`{u} F, Ranon R, Christie M (2018) Thinking like a director: Film
  editing patterns for virtual cinematographic storytelling. ACM Transactions
  on Multimedia Computing, Communications, and Applications 14(4),
  \doi{10.1145/3241057}

\bibitem[{Yu and Lisin(2009)}]{Yu2009}
Yu SX, Lisin DA (2009) Image compression based on visual saliency at individual
  scales. In: Advances in Visual Computing (ISVC), pp 157--166,
  \doi{10.1007/978-3-642-10331-5_15}

\bibitem[{Zhang and Chen(2019)}]{kao2019model}
Zhang K, Chen Z (2019) Video saliency prediction based on spatial-temporal
  two-stream network. IEEE Transactions on Circuits and Systems for Video
  Technology 29(12):3544--3557, \doi{10.1109/TCSVT.2018.2883305}

\bibitem[{Zhang et~al.(2020)Zhang, Saran, Liu, Zhu, Guo, Niekum, Ballard, and
  Hayhoe}]{Zhang2020}
Zhang R, Saran A, Liu B, Zhu Y, Guo S, Niekum S, Ballard D, Hayhoe M (2020)
  Human gaze assisted artificial intelligence: A review. In: Proceedings of the
  Twenty-Ninth International Joint Conference on Artificial Intelligence,
  IJCAI-20, pp 4951--4958, \doi{10.24963/ijcai.2020/689}

\bibitem[{Z{\"u}nd et~al.(2013)Z{\"u}nd, Pritch, Sorkine-Hornung, Mangold, and
  Gross}]{Zund2013}
Z{\"u}nd F, Pritch Y, Sorkine-Hornung A, Mangold S, Gross T (2013)
  Content-aware compression using saliency-driven image retargeting. 2013 IEEE
  International Conference on Image Processing (ICIP) pp 1845--1849,
  \doi{10.1109/ICIP.2013.6738380}

\end{thebibliography}
\end{document}